\documentclass[journal]{IEEEtran}
\usepackage{xcolor,soul,framed}
\usepackage[noadjust]{cite}
\usepackage{tikz}

\usepackage{pgfplots}
\usepackage{glossaries}
\usepackage{float}
\usepackage{rotating,multirow}
\usepackage{acronym}

\colorlet{shadecolor}{yellow}
\graphicspath{{../pdf/}{../jpeg/}}
\DeclareGraphicsExtensions{.pdf,.jpeg,.png}

\usepackage{array}
\usepackage{mdwmath}
\usepackage{mdwtab}
\usepackage{eqparbox}
\usepackage{url}
\usepackage{caption}
\usepackage{subcaption}

\usepackage{amsmath,amsfonts, amssymb}
\usepackage{mathtools}
\usepackage{acronym}
\usepackage{longtable}
\usepackage{textcomp}
\usepackage{pifont}
\usepackage{graphicx}
\usepackage{color}
\usepackage{epsfig}
\usepackage{url}
\usepackage{balance}
\usepackage{float}
\usepackage{array}
\usepackage{comment}
\usepackage{siunitx}
\usepackage{enumerate}   
\usepackage{todonotes}
\usepackage{xr}
\usepackage{caption}
\usepackage{subcaption}
\usepackage{lineno,hyperref}
\usepackage{multirow}
\usepackage{adjustbox}
\usepackage{booktabs}
\usepackage{xcolor}
\usepackage{amsthm}
\usepackage{soul}
\usepackage{tcolorbox}
\usepackage{fancyvrb}

\usepackage{algorithm}
\usepackage{algpseudocode}
\usepackage{xr}

\usepackage{rotating}
\usepackage{tabularray}

\makeatletter
\newcommand*{\addFileDependency}[1]{
  \typeout{(#1)}
  \@addtofilelist{#1}
  \IfFileExists{#1}{}{\typeout{No file #1.}}
}
\makeatother

\def\BibTeX{{\rm B\kern-.05em{\sc i\kern-.025em b}\kern-.08em
    T\kern-.1667em\lower.7ex\hbox{E}\kern-.125emX}}
\AtBeginDocument{\definecolor{ojcolor}{cmyk}{0.93,0.59,0.15,0.02}}
\def\OJlogo{\vspace{-14pt}\includegraphics[height=28pt]{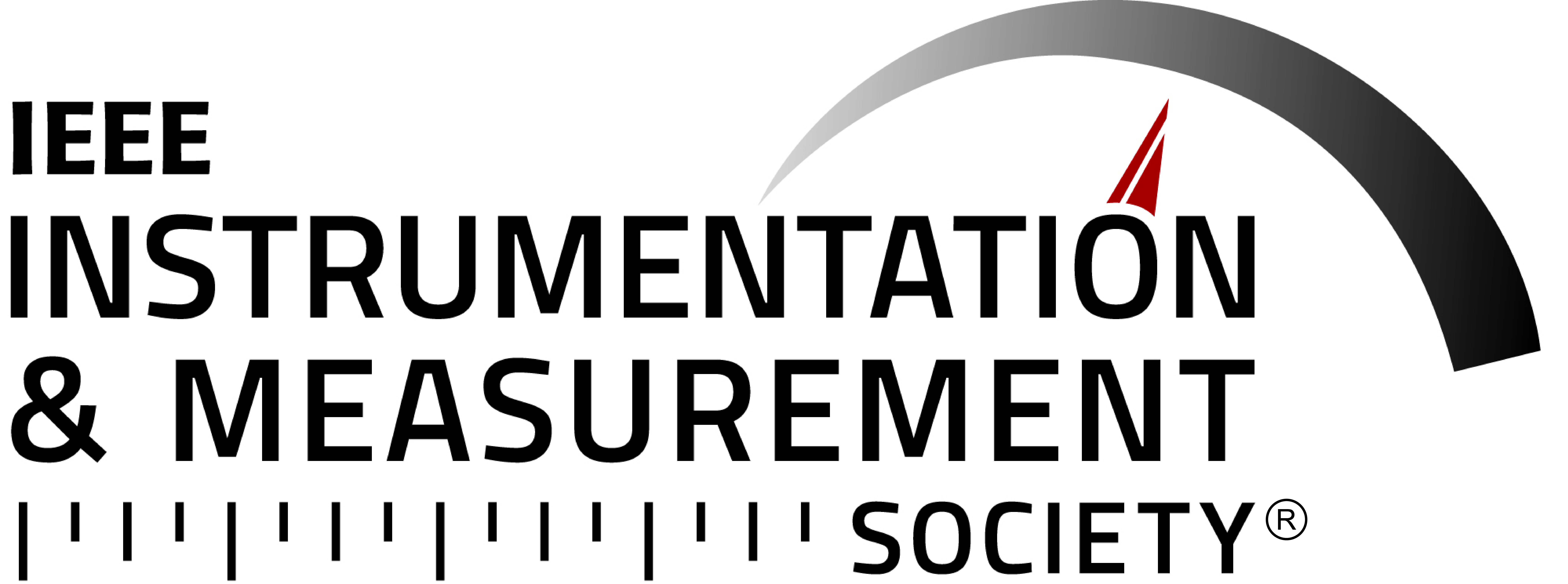}}

\def\@IEEEauthorblockNstyle{\normalfont\@IEEEcompsocnotconfonly{\sffamily}\sublargesize\@IEEEcompsocconfonly{\large}}
\def\@IEEEauthorblockAstyle{\normalfont\@IEEEcompsocnotconfonly{\sffamily}\@IEEEcompsocconfonly{\itshape}\normalsize\@IEEEcompsocconfonly{\large}}

\newcommand{\cmark}{\ding{51}}
\newcommand{\xmark}{\ding{53}}

\IEEEoverridecommandlockouts

\newcommand{\proto}{URANUS}

\newcommand{\R}{\mathbb{R}}
\newcommand{\N}{\mathbb{N}}

\acrodef{CAGR}{Compound Annual Growth Rate}
\acrodef{DT}[DT]{Decision Tree}
\acrodef{CSL}{Coordinated Sampled Listening}
\acrodef{AES}{Advanced Encryption Standard}
\acrodef{SDR}{Software-Defined Radio}
\acrodef{SFD}{Start of Frame Delimiter}
\acrodef{MTU}{Maximum Transmission Unit}
\acrodef{CRC}{Cyclic Redundancy Check}
\acrodef{ASN}{Absolute Slot Number}
\acrodef{LPW}{Low Power Networks}
\acrodef{MAC}{Medium Access Control}
\acrodef{RR}[RR]{Radar Reflectivity}
\acrodef{DNN}[DNN]{Deep Neural Network}
\acrodef{CNN}[CNN]{Convolutional Neural Network}
\acrodef{DRNN}[DRNN]{Deep Recurrent Neural Network}
\acrodef{GoF}[GoF]{Goodness-of-Fit}
\acrodef{YOLO-lite}[YOLO-lite]{You Only Look Once - lite}
\acrodef{XGBoost}[XGBoost]{eXtreme Gradient Boosting}
\acrodef{AWGN}[AWGN]{Additive white Gaussian noise}
\acrodef{CAM}{Class Activation Mapping}
\acrodef{PSD}{Power Spectrum Density}
\acrodef{UAV}{Unmanned Aerial Vehicle}
\acrodef{C-UAS}{Counter Unmanned Aerial System}
\acrodef{MCAR}{Missing completely at Random}
\acrodef{MAR}{Missing at Random}
\acrodef{MNAR}{Missing not at Random}
\acrodef{GMM}{Gaussian Mixture Models}
\acrodef{ANOVA}{Analysis of Variance}
\acrodef{NATO}{North Atlantic Treaty Organization}
\acrodef{RCS}{Radar cross-section}
\acrodef{FCSF}{Frame Cross Section Frontal}
\acrodef{NDZ}{No-Drone Zone}
\acrodef{RSS}{Receiving Signal Strength}
\acrodef{FAA}{Federal Aviation Administration}
\acrodef{FC}{fully connected}
\acrodef{FC-DNN}{Fully Connected Deep Neural Networks}
\acrodef{KNN}{K-Nearest Neighbor}
\acrodef{SNR}{Signal-to-Noise Ratio}
\acrodef{FD}{Fractal Dimension}
\acrodef{AIB}{Axially Integrated Bispectra}
\acrodef{SIB}{Square Integrated Bispectra}
\acrodef{PCA}{Principal Component Analysis}
\acrodef{NCA}{Neighborhood Component Analysis}
\acrodef{SVM}{Support Vector Machine}
\acrodef{MLP}{Multilayer Perceptron}
\acrodef{DT}{Decision Tree}
\acrodef{AI}{Artificial Intelligence}
\acrodef{ML}{Machine Learning}
\acrodef{DL}{Deep Learning}
\acrodef{GIS}{Geographic Information System}
\acrodef{UTM}{Universal Transverse Mercator}
\acrodef{GCS}{Geographic Coordinate System}
\acrodef{DMS}{Degrees Minutes Seconds}
\acrodef{DD}{Decimal Degree}
\acrodef{RFDF}{Radio Frequency / Direction Finding}
\acrodef{PDF}{Probability Density Function}
\acrodef{MAE}{Mean Absolute Error}
\acrodef{MSE}{Mean Squared Error}
\acrodef{SSE}{Sum of Squared Errors}
\acrodef{FLP}[FLP]{Flight Pattern}
\acrodef{CSV}{Comma-separated Values}
\acrodef{DL}{Deep Learning}
\acrodef{RF}[RF]{Random Forest}
\acrodef{ANN}[ANN]{Artificial Neural Network}
\acrodef{NATO}[NATO]{North Atlantic Treaty Organization}
\acrodef{CI}[CI]{Critical Infrastructure}
\acrodef{ReLU}[ReLU]{Rectified Linear Unit}
\acrodef{GUI}[GUI]{Graphical User Interface}
\acrodef{SIM}[SIM]{Subscriber Identity Module}
\acrodef{LTE}[LTE]{Long Term Evolution}
\acrodef{TP}{True Positive}
\acrodef{FP}{False Positive}
\acrodef{FN}{False Negative}

\begin{document}

\title{URANUS: Radio Frequency Tracking, Classification and Identification of Unmanned Aircraft Vehicles}

\author{
\IEEEauthorblockN{
Domenico Lofù\IEEEauthorrefmark{1},
Pietro Di Gennaro\IEEEauthorrefmark{2},
Pietro Tedeschi\IEEEauthorrefmark{3}, 
Tommaso Di Noia\IEEEauthorrefmark{1}, and
Eugenio Di Sciascio\IEEEauthorrefmark{1}
}

\IEEEauthorblockA{
\IEEEauthorrefmark{1}Dept. of Electrical and Information Engineering (DEI), Politecnico di Bari, Bari, Italy\\ \{domenico.lofu, tommaso.dinoia, eugenio.disciascio\}@poliba.it}\\
\IEEEauthorblockA{\IEEEauthorrefmark{2}IMT School for Advanced Studies Lucca, Lucca, Italy}\\
pietro.digennaro@imtlucca.it\\
\IEEEauthorblockA{\IEEEauthorrefmark{3}Autonomous Robotics Research Center, Technology Innovation Institute, Abu Dhabi, United Arab Emirates\\
pietro.tedeschi@tii.ae}\\
\thanks{This is a personal copy of the authors. Not for redistribution. The final published version of the paper accepted in IEEE Open Journal of Vehicular Technology  Journal is available through the IEEExplore Digital Library, with the DOI: 10.1109/OJVT.2023.3333676.}
}

\maketitle

\begin{abstract}
Safety and security issues for Critical Infrastructures are growing as attackers adopt drones as an attack vector flying in sensitive airspaces, such as airports, military bases, city centers, and crowded places. Despite the use of UAVs for logistics, shipping recreation activities, and commercial applications, their usage poses severe concerns to operators due to the violations and the invasions of the restricted airspaces. A cost-effective and real-time framework is needed to detect the presence of drones in such cases. In this contribution, we propose an efficient radio frequency-based detection framework called \proto. We leverage real-time data provided by the Radio Frequency/Direction Finding system, and radars in order to detect, classify and identify drones (multi-copter and fixed-wings) invading no-drone zones. We adopt a Multilayer Perceptron neural network to identify and classify UAVs in real-time, with $90$\% accuracy. For the tracking task, we use a Random Forest model to predict the position of a drone with an MSE $\approx0.29$, MAE $\approx0.04$, and $R^2\approx 0.93$. Furthermore, coordinate regression is performed using Universal Transverse Mercator coordinates to ensure high accuracy. Our analysis shows that \proto\ is an ideal framework for identifying, classifying, and tracking UAVs that most Critical Infrastructure operators can adopt.
\end{abstract}


\section{Introduction and Motivation}
\label{sec:IntroMotivation}
\IEEEPARstart{I}{}n the last years, \acp{UAV} commonly known as \emph{drones} have become a crucial technology for several types of applications such as environmental monitoring~\cite{Liao2021_IOTM, Bacco2018_ICL, Inoue2019_IGARSS}, smart grids control \cite{Zhou2018_ITII}, crime prevention~\cite{Zheng2020_ITEC, Huang2022_ITASE}, smart cities management~\cite{Alsamhi2019_DRONES}, and the military operations~\cite{dipietro2019_jamme}. According to authoritative marketing research industries, the \ac{UAV} market is estimated to be $500,000$ units in $2025$ and is projected to reach $6.9$ million units by $2030$. The global drone logistics \& transportation market accounted for US\$ $24.58$ million in 2018 and is expected to grow at a \ac{CAGR} of $60.6$\% over the forecast period $2019-2027$, to account for US\$ $1,626.98$ million in $2027$~\cite{marketdrone}. Factors including increasing developments in the e-commerce sector~\cite{Lee2022_ITVT} and rising acceptance owing to various benefits offered are significantly driving the global drone logistics~\cite{song2018persistent} \& transportation market~\cite{moshref2021applications}.

Most commercial drones are autonomous or remotely controlled vehicles, that leverage the standard Wi-Fi frequency bands, \emph{i.e.} $2.4$~\emph{GHz}~\cite{fabra2017impact} and $5.0$~\emph{GHz}~\cite{zeng2018cellular}. They can be programmed to execute tasks that span from object tracking~\cite{fu2023siamese}, and delivery, to committing illegal activities such as privacy violations, destroying critical infrastructures, and harming public safety during crowded events~\cite{tedeschi_tdsc2022}. Given the above threats, several countermeasures~\cite{yaacoub2020security} based on audio, video, thermal, and Radio Frequency signals have been exploited in the last few years for drone identification and tracking. However, the performance of these systems is affected when the surrounding environment is impaired (e.g. weather conditions, noise, low light visibility). Indeed, most critical infrastructures adopt \ac{RFDF} and kinematics radar sensors that track all types of drones by analyzing the reflected signals and comparing them to a database for drone characterization. Due to the high number of unauthorized \acp{UAV} operating in the skies, it is crucial to deploy a system framework to track, classify and identify timely, malicious \acp{UAV} by leveraging the data provided by radar sensors.

In this paper, we design \proto, a \ac{ML} framework that analyzes a dataset with data extracted from 
two \ac{RFDF} sensors namely \textit{Diana} and \textit{Venus}, and two radar sensors namely \textit{Arcus} and \textit{Alvira} to \textit{(i) identify}, \textit{(ii) classify}, and \textit{(iii) track} drone(s) on a \ac{NATO} military base (placed in the \ac{C-UAS} test centre in the Netherlands).

Our framework is trained over a real dataset derived from a data source of \acp{UAV} flights provided by \ac{NATO}~\cite{Kaggle2021_DATNATO}.

Our prototype adopts popular libraries and tools such as \textit{PyTorch}, \textit{scikit-learn} and \textit{pandas}, available online as open source code~\cite{uranus_code}.

Our main contributions include the following:
\begin{enumerate}
    \item Identify, classification, and tracking of one or more flying drones;
    \item Classification of fixed-wing and multi-copter drones;
    \item Analysis of both \ac{RFDF}, and kinematics sensors to detect drones in \ac{CI};
    \item Real-time framework execution.
\end{enumerate}

The remainder of this paper is organized as follows: Section~\ref{sec:Preliminaries} introduces the technical background. Section~\ref{sec:scenario_adv} describes the reference scenario and adversarial model, while Section~\ref{sec:DataSetAnalysis} details the dataset analysis.
Section~\ref{sec:PropArchitecture} shows the proposed architecture, while experiments and results are described in Section~\ref{sec:ExperimentsAndResults}. Section~\ref{sec:RelatedWork} reviews the related works, and Section~\ref{sec:Conclusion} concludes the paper.

\section{Preliminaries}
\label{sec:Preliminaries}
In this section, we introduce some preliminary knowledge adopted throughout the rest of the manuscript.
Specifically, section~\ref{sec:GCS and UTM} describes the coordinate systems adopted in this work, while section~\ref{sec:RadarParam} and~\ref{sec:ML_AlgoAdopted} describes the radar parameters and the \ac{ML} models used in the \proto, respectively.

\subsection{GCS and UTM Coordinate Systems}
\label{sec:GCS and UTM}
The \ac{GCS}~\cite{lu2004global} and the \ac{UTM} coordinate system~\cite{langley1998utm} are two standard techniques to represent locations on the Earth's surface. Coordinates systems often use a tuple of real numbers $(x_1, x_2)$ to identify an object's unique location, where $x_1 \in \R$, and $x_2 \in \R$.
\\\\
\textbf{Geographic Coordinate System (GCS)}. Latitude and longitude are used to specify the location of a point on the Earth's surface. The \emph{latitude} (Eq.~\ref{eq:latitude}) measures the distance north or south of the Equator, while the \emph{Longitude} (Eq.~\ref{eq:longitude}) measures the distance east or west of the Prime Meridian:

\begin{equation}
\label{eq:latitude}
    \text{Latitude} = \arcsin\left(\frac{Z}{\sqrt{X^2 + Y^2 + Z^2}}\right),
\end{equation}
\begin{equation}
\label{eq:longitude}
    \text{Longitude} = \arctan\left(\frac{Y}{X}\right),
\end{equation}
where $X$, $Y$, and $Z$, represent the Cartesian coordinates of the object to track. Longitude and latitude are defined in the \ac{DMS} form or using \acp{DD} values. An example with coordinates expressed in \ac{DMS} format is:
\begin{center}
    ($-73^{\circ}$ $58'$  $2"$, $40^{\circ}$ $44'$ $58"$)\\
\end{center}
while the same place expressed in \ac{DD} format is:
\begin{center}
    ($-73.967385$, $40.749598$).
\end{center}
\noindent
\textbf{Universal Transverse Mercator (UTM)}. The Earth is segmented into $60$ longitudinal zones, each spanning $6$ degrees of longitude~\cite{Inggs2009_IRC}. Within each zone, a transverse Mercator projection is used to represent locations:
\begin{equation}
    \text{Easting} = \text{Zone Number} \times 10^5 + \text{Easting Value},
\end{equation}
\begin{equation}
    \text{Northing} = \text{Northern Hemisphere Constant} + \text{Northing Value},
\end{equation}
where \emph{Zone Number}, \emph{Easting Value}, and \emph{Northing Value} represent the longitudinal zone, the distance east of the central meridian in \emph{meters}, and the distance north of the Equator in \emph{meters}, respectively. Further, the Northern Hemisphere has a constant of $0$, and the Southern Hemisphere has a value of $10,000,000$~\cite{Inggs2009_IRC}. The Zone Number and the Hemisphere are adopted to uniquely identify locations within the \ac{UTM} grid. An example of coordinates in \ac{UTM} is:
\begin{center}
    $18\ N\ 587173\ 4511473$
\end{center}
where $18$ in the \ac{UTM} zone, $N$ is the Northern Hemisphere, $587173$ is the Easting value and $4511473$ is the Northing value.

To the best of our knowledge, we are the first to highlight better performances of an \ac{ML} model that adopts \ac{UTM} coordinates instead of the canonical \ac{GCS}. In our regression tests with \ac{RF} models, the mean difference between real and predicted values is around $18$ meters, a relevant value. 
Furthermore, the minimum margin between real values and positions provided by Radar Sensor Systems was set to $50$ meters by project settings.

\subsection{Range and Bearing Radar Parameters}
\label{sec:RadarParam}
In radar systems, \emph{Range} and \emph{Bearing} are two fundamental concepts used to determine the location of a target.\\\\
\noindent
\textbf{Range}. The range refers to the distance between the radar system and the target. It represents the radial distance from the radar transmitter/receiver to the reflecting point on the target. The range (Eq.~\ref{eq:range_corr}) is typically measured in units such as \emph{meters} or \emph{nautical miles}, as follows: 
\begin{equation}\label{eq:range_corr}
    R = \frac{c\cdot T_R}{2},
\end{equation}
where $c = 3\cdot10^8 m/s$ is the speed of the light, and $T_R$ is the transmitted pulse~\cite{Blair1998_ITAES}.\\\\
\noindent
\textbf{Bearing}. In radar systems, bearing refers to determining the direction from which a detected signal or echo comes. The True Bearing, referenced to true north, for a radar target is the angle formed between true north and a line directly aimed at the target~\cite{lara2023_irs}. Radar systems determine bearing by analyzing the angle at which the received signal is stronger. This latter is usually achieved by using an antenna array or rotating the antenna to scan the surrounding environment.

\subsection{Machine Learning Algorithms}
\label{sec:ML_AlgoAdopted}
This section introduces the \ac{ML} algorithms adopted in our framework.
\\\\
\textbf{Multilayer Perceptron (MLP)}. 
\ac{MLP}, also known as \textit{Deep Feedforward Networks} or \textit{Feedforward Neural Networks}~\cite{goodfellow}, it is a type of \ac{ANN}~\cite{Lin2022_ITNNLS}\cite{goodfellow} widely used in Machine Learning and pattern recognition tasks. 
Specifically, we adopted this technique for the classification task.
A \ac{MLP} mimics how neurons interact and work in the human brain. It is characterized by a layered architecture consisting of multiple interconnected nodes organized into three primary layers: the input layer, one or more hidden layer(s), and the output layer. The hidden and output layers have neurons connected to the neurons of their preceding layers and network connections; further, the topology can be fully connected or partially connected.\\
In MLP neural networks, each unit performs a biased, weighted sum of inputs and passes this activation level through a transfer function to generate output. The \ac{ReLU} activation function is the preferred default activation function for most feed-forward neural networks. When applied to the output of a linear transformation, it results in a nonlinear transformation.\\
\acp{MLP} have demonstrated outstanding capabilities in modelling complex data patterns for several deep learning architectures. They can be customized with several hidden layers and neurons to accommodate the complexity of different tasks. Each unit resembles a neuron, i.e., it receives input from many other units and computes its activation value.\\
By tuning the weights of the connections between the nodes in the network, the model learns to predict the target output.
Further, an optimization algorithm is adopted to adjust the weights (\emph{i.e. stochastic gradient descent}).
In particular, it minimises the difference between the predicted and the actual target output.\\
The \ac{MLP} model is represented as a function $f(x)$ that maps the input data $x$ to the output $y$. The function $f(x)$ is expressed as a composition of other functions, as shown in the following equation:
\begin{equation}
    f_a(x)=f_L(f_{L-1}(...(f_2(f_1(x,\theta_1),\theta_2)...) ,\theta_{L-1}),\theta_L) 
\end{equation}
where $f_i$ is the nonlinear transfer function of the $i^{th}$ hidden layer, $\theta_i$ represents the weights connecting the nodes in layer $i$ and layer $i+1$, and $L$ is the number of layers in the \ac{MLP}.
\\\\
\textbf{Random Forest (RF)}. It is a typical \ac{ML} algorithm~\cite{random_forest2},\cite{random_forest3} used for the regression task. 
A \ac{RF} model is an ensemble of decision trees that can handle high-dimensionality datasets. Each decision tree is trained on a subset of the data and a subset of the features. The output of the individual decision tree is the average or mode for regression or classification, respectively.

Let \ac{SSE} be the sum of the squared differences between the predicted and actual values of the target variable. In the regression form, the splitting criteria for node creation follows the largest reduction in the value of the \ac{SSE} for the predicted output. The splitting process continues until a stopping criteria (such as a maximum tree depth or a minimum number of data points in a leaf node) is defined. The output of the \ac{RF} for regression is the average of the predicted values of the individual decision tree. The output value of a \ac{RF} for regression is:
\begin{equation}
    \hat{y} = \frac{1}{M} \sum_{i=1}^{M} f_i(x)
\end{equation}
where $\hat{y}$ is the predicted value of the target variable, $M$ is the number of decision trees in the \ac{RF} model, and $f_i(x)$ are the predictions of the $i-th$ decision tree for the input features. We adopt the Random Forest in \proto\ for the coordinates regression task.

\subsection{Preprocessing Primitives}
\label{sec:preprocessing_prims}
This section introduces the preprocessing primitives used to transform the original data for the \ac{ML} models~\cite{Hutter2019_AUTOMATED}.
\\\\
\textbf{Dataset Standardization}. It is an essential preprocessing step in \ac{ML}, suitable for algorithms sensitive to the scale of input features~\cite{standardization2}. In order to identify patterns and relationships with high accuracy, input features should have the same scale. Input data should be adapted to have $0$ mean and a standard deviation equal to $1$. To this aim we apply Eq.~\ref{eq:dataset_std} to every single feature of the initial dataset:
\begin{equation}\label{eq:dataset_std}
    x_{sf} = \frac{x - \mu}{\sigma}
\end{equation}
where, $x$ is the considered feature value, $\mu$ and $\sigma$ represents the mean and the standard deviation of the feature, and $x_{sf}$ is the final scaled feature value. 
\\\\
\textbf{One-hot Encoding}. It is a widely used technique~\cite{Cerda2022_ITKDE, rodriguez2018beyond} for converting categorical data into a binary matrix format suitable for \ac{ML} models. Different algorithms require numerical input, and the one-hot encoding makes the representation of discrete categories as unique binary vectors.\\
Specifically, the~Eq.~\ref{eq:onehotenc} shows how this technique encodes categories:

\begin{equation}
\label{eq:onehotenc}
    b_{i,j} = \begin{cases} 1 &\quad\text{if } i=j\\
    0 &\quad\text{otherwise} \end{cases}
\end{equation}

The vector and category index are represented by $i$ and $j$, respectively, while the output is a binary vector namely $\bar{b}$.
For any given observation, the binary vector $\bar{b}$ column corresponding to its category is marked with a $1$, while all other columns are set to $0$. This representation ensures the avoidance of incorrect relational order between categories.
\\\\
\textbf{Label encoding}. This widely used technique converts categorical data into numerical form~\cite{goodfellow} and makes it machine-readable, as algorithms work exclusively with numerical data. In this approach, each unique category or label within a feature is assigned a distinct integer value. For instance, for a categorical variable with values like \textit{low}, \textit{medium}, and \textit{high}, label encoding might assign these categories values of $0$, $1$, and $2$, respectively. The primary advantage of label encoding is its simplicity and ability to retain a compact dataset representation.

\section{Reference Scenario and Adversarial Model}
\label{sec:scenario_adv}
This section introduces the scenario and the adversarial model considered in our work. Specifically, section~\ref{sec:cuas_scenario} depicts the system model and describes the assumptions, while section~\ref{sec:adv_model} describes the adversary model.

\subsection{System Model and Assumptions}
\label{sec:cuas_scenario}
\begin{figure}[htbp]
    \centering
    \includegraphics[width=1\columnwidth]{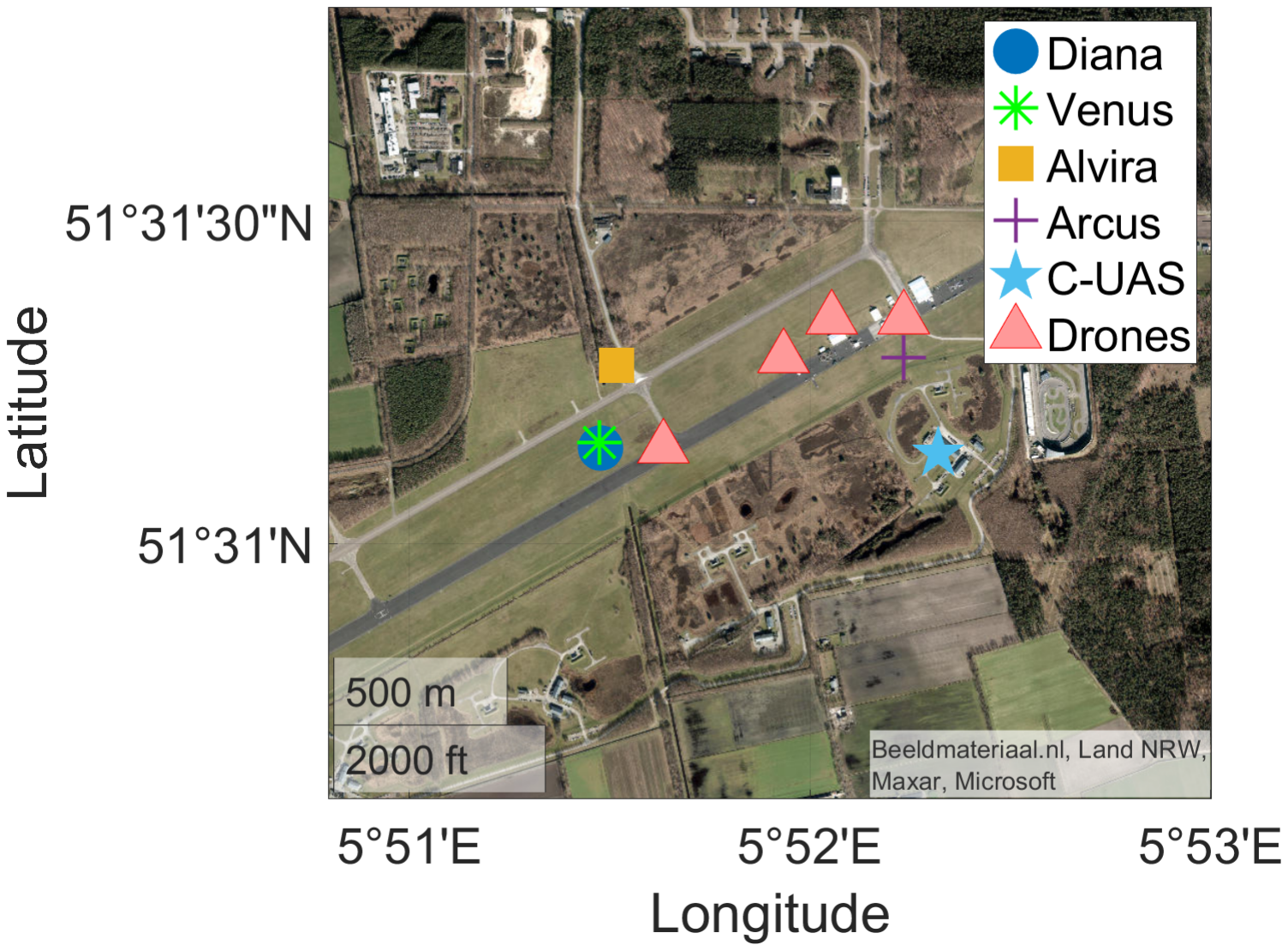}
    \caption{Scenario assumed in this work.}
    \label{fig:scenario}
\end{figure}

The scenario assumed in this work is depicted in Figure~\ref{fig:scenario}. We consider the problem of tracking, classifying, and identifying one or multiple drones (multi-copter or fixed wing) in a No-Drone Zone~\cite{faa_ndz}. It is worth noticing that we refer to the scenario described by \ac{NATO} in~\cite{Kaggle2021_DATNATO}.

The main components of the considered scenario are the following:
\begin{itemize}
    \item \ac{NDZ} - The \ac{FAA} introduced the term \ac{NDZ} to describe an area that does not allow to operate by using a drone or unmanned aircraft system. Examples of \ac{NDZ} areas are airports, restricted airspaces, government agencies, or also temporary flight restrictions areas such as sports events, presidential movements, and security-sensitive areas.
    
    \item Drone - The drone(s) assumed in this scenario are classified with the features reported in Table~\ref{tab:drone_classification}. Each drone is characterized by the commercial name, a code field, and a unique identifier assigned during the drone configuration phase. The airframe represents the type of drone, i.e. a multi-copter or a fixed-wing drone; the weight specifies the maximum weight of the drone, and velocity indicates the maximum speed of the drone. \ac{RCS} or radar signature measures how much energy the drone reflects towards a radar, i.e. the area seen by radar. \ac{FCSF} defines a frame of the frontal measurement of the cross-section.
    \begin{table}[htbp]
    \centering
    \caption{Classification and characteristics of the drones.}
    \label{tab:drone_classification}
    \resizebox{\columnwidth}{!}{
        \begin{tabular}{cccccc}
            \toprule
            Drone Name        & Airframe    & Weight   & Max Velocity & RCS      & FCSF     \\
                              &             & $[kg]$ & $[m/s]$    & $[m^2]$ & $[m^2]$ \\ \hline
            DJI Mavic Pro     & Multi-copter & $1$        & $20$           & $0.01$     & $0.02$     \\
            DJI Mavic 2       & Multi-copter & $1$        & $20$           & $0.01$     & $0.02$     \\
            DJI Phantom 4 Pro & Multi-copter & $1$        & $20$           & $0.01$     & $0.02$     \\
            Parrot Disco      & Fixed wing  & $1$        & $20$           & $0.005$    & $0.1$      \\ 
            \bottomrule
        \end{tabular}
    }
    \end{table}
    
    \item RFDF-Radar Sensor Network - The \acl{NDZ} is monitored by two \ac{RFDF} sensors, namely Diana and Venus, and two radar sensors i.e. Arcus and Alvira (all of them are fictitious names). From one side, Diana and Venus sensors acquire data such as time-of-arrival (timestamp), \ac{RSS}, and beamforming to localize the target. Diana adopts a linear array antenna to estimate the bearing of an intercept; it only reports detections in a $180^\circ$ sector even if the target is located in the opposite sector. Venus uses a circular array antenna with no bearing ambiguity and provides no range information. Conversely, Arcus and Alvira sensors are 2D radar and 3D radar, respectively. These sensors provide crucial information such as latitude, longitude, altitude, and timestamp, as well as the bearing and range of the drone during the flight. Table~\ref{tab:sensors} summarizes the name, the type, and the sensor location (latitude and longitude coordinates).
    \begin{table}[htbp]
    \centering
        \caption{Description of the sensor types with their respective positions.}
        \label{tab:sensors}
        \begin{tabular}{cccc}
            \toprule
            Sensor Name &  Sensor Type &  Sensor Latitude &  Sensor Longitude \\ \hline
            Diana & RFDF & $51.51913^\circ$ & $5.85795^\circ$\\
            Venus & RFDF & $51.51927^\circ$ & $5.85791^\circ$ \\
            Alvira & Radar & $51.52126^\circ$ & $5.85860^\circ$ \\
            Arcus & Radar & $51.52147^\circ$ & $5.87056^\circ$ \\ 
            \bottomrule
        \end{tabular}
    \end{table}
     
    \item \acl{C-UAS} - It is a central server unit adopted to collect and process (via \proto) the data generated by the sensor network. The system is used to detect, identify and track the presence of any unauthorized or malicious drone in the \acl{NDZ}.
\end{itemize}

As mentioned above, our scenario assumes the presence of a \ac{C-UAS} operator in a \ac{NDZ}, \emph{e.g.,} the one controlling a generic critical infrastructure. Such operators are interested in monitoring the critical infrastructure, looking for malicious and unauthorized \acp{UAV} approaching a sensitive area. To this aim, the operators adopt an \ac{RFDF} Radar sensor network to capture crucial information in the monitored area to identify, classify and track the owner of the \ac{UAV}. In this paper, we consider three macro-scenarios as follows:
\begin{itemize}
    \item Scenario 1.1, 1.2, 1.3, and 1.4. In these scenarios, we assume a single \ac{UAV} (multi-copter) is flying in the \ac{NDZ}.
    \item Scenario 2.1 and 2.2. In these scenarios, we assume two \acp{UAV} (multi-copter) are flying in the \ac{NDZ}.
    \item Scenario 3. In this scenario, we assume one \acp{UAV} (fixed-wing) is flying in the \ac{NDZ}.
\end{itemize}

Each flight scenario involves one or more drones with individual flight patterns. According to Table~\ref{tab:scenario_drone}, the various scenarios include different drones.

\begin{table}[!htbp]
\centering
\caption{Drone(s) model involved in each scenario.}\label{tab:scenario_drone}
\begin{tabular}{cc}
    \toprule
    Scenario & Drone Name                 \\ \hline
    Scenario 1.1      & DJI Mavic Pro                       \\
    Scenario 1.2      & DJI Phantom 4 Pro                   \\
    Scenario 1.3      & DJI Mavic Pro                       \\
    Scenario 1.4      & DJI Mavic 2                       \\
    Scenario 2.1      & DJI Phantom 4 Pro and DJI Mavic 2   \\
    Scenario 2.2      & DJI Phantom 4 Pro and DJI Mavic Pro \\
    Scenario 3        & Parrot Disco                        \\ 
    \bottomrule
    \end{tabular}
\end{table}

We highlight that the aforementioned scenarios are only a reference for the considered dataset. Our framework can be applied to other potential environments, such as surveillance towers in critical infrastructures, military bases, ports, and airports.

\subsection{Adversary Model}
\label{sec:adv_model}
In all the scenarios, we assume that an adversary $\mathcal{E}$ has the capabilities to radio-control a single drone or a swarm of drones, and it is interested in reaching a target inside a \ac{NDZ}. The aims of the adversary can be manifold, \emph{e.g.,} violating the privacy of the area by recording video and/or taking photos of a sensitive area, using it as a bomb in critical infrastructures such as airports, oil\&gas industries, nuclear power plants, water treatment facilities, ports, telecommunication networks, or to threat people safety by carrying explosives or radioactive materials or colliding with airplanes during the take-off and landing procedures. Moreover, the adversary $\mathcal{E}$ can control a drone in several ways: (i) through a wireless remote controller, (ii) remotely via the Internet (i.e. the drone supports embedded \ac{SIM}, standard \ac{SIM}, and cellular \ac{LTE} or 5G technology), (iii) by pre-programming it through way-points to enable the autonomous flight. Conversely, we assume that the drones broadcast data (for several purposes) via the onboard radio transmitter for the whole flight~\cite{tedeschi2021_acsac}.

\section{Dataset Analysis}
\label{sec:DataSetAnalysis}
In this section, we describe the data source used to develop the~\proto\ framework. Further, we outline the challenges and the proposed techniques to make the dataset suitable for the training and testing phases for our \ac{ML} models.
\\\\
\textbf{Dataset Preprocessing}. Preparing raw data for \ac{ML} analysis and modelling is a critical step. This step is crucial to guarantee that the data are (i) consistent, (ii) flawless, and (iii) complete. These characteristics allow \ac{ML} models to learn efficiently from the data and make highly accurate predictions. We considered a data source provided by \ac{NATO} containing real \ac{UAV} flight measurements recorded in 2020.

The data source is available online as \ac{CSV} format~\cite{Kaggle2021_DATNATO} which contains $65$ files ($366$ \emph{MBs}) organized in two main sub-folders, namely \emph{training} and \emph{test} folder.
 
Due to the lack of label information, the test data folder is not considered in our work. We leveraged only the training folder data that contains $37$ files ($194$ \emph{MBs}) organized in seven scenarios. In detail, for each scenario, we have data related to (i) radar and \ac{RFDF} systems sensor and (ii) \ac{UAV} flight parameters. Specifically, Table~\ref{table:data_log} and Table~\ref{table:data_sensor} report data examples from \emph{log} and \emph{sensor data} files. It is worth noticing that for every single drone, the log file records \acp{UAV} parameters such as the \emph{timestamp}, \emph{latitude}, \emph{longitude}, \emph{speed}, \emph{altitude}, and \emph{drone type}.

Furthermore, the sensor data features such as the \ac{RCS}, \ac{RF}, and the \ac{UAV} parameters are stored along with the related timestamp. Accordingly, the timestamp is adopted as the index of each data sample to merge and correlate the sensor readings and the drone data logs.
\\\\

\begin{table*}[t]
\centering
\caption{Data sample from drone log of scenario 1.1.}
\label{table:data_log}
	\hspace*{-0.4cm}
	\begin{tabular}{cccccccc}
        \toprule	
		Latitude & Longitude & Altitude [$m$] & UltrasonicHeight [$m$] & Speed [$m/s$] & Distance [$m$] from radar & Datetime [UTC] & Timestamp \\
		\hline
        $51.519506$ & $5.857978$ & $0.9$ &  $0$  & $2.3$ & $0.04$ & 2020-09-29 12:10:56.647 & $1601381456647$ \\
		$51.519506$ & $5.857978$ & $1.2$ & $1.2$ & $2.7$ & $0.05$ & 2020-09-29 12:10:56.727 & $1601381456727$ \\
		$51.519506$ & $5.857978$ & $1.5$ & $1.5$ & $3.1$ & $0.06$ & 2020-09-29 12:10:56.824 & $1601381456823$ \\
		$51.519506$ & $5.857978$ & $1.8$ & $1.9$ & $3.5$ & $0.06$ & 2020-09-29 12:10:56.928 & $1601381456927$ \\
		$51.519506$ & $5.857978$ & $2.2$ & $2.2$ & $3.7$ & $0.07$ & 2020-09-29 12:10:57.027 & $1601381457027$ \\
		$51.519506$ & $5.857978$ & $2.6$ & $2.6$ & $3.8$ & $0.08$ & 2020-09-29 12:10:57.126 & $1601381457126$ \\
		$51.519506$ & $5.857979$ & $2.9$ & $3$   & $3.7$ & $0.09$ & 2020-09-29 12:10:57.225 & $1601381457224$ \\
		\bottomrule
	\end{tabular}
\end{table*}

\begin{table*}[t]
\centering
\caption{Data sample from radar sensor ARCUS, in scenario 1.1.}
\label{table:data_sensor}
	\hspace*{0.7cm}
	\begin{tabular}{cccccccc}
        \toprule	
		Timestamp & Latitude & Longitude & Altitude [$m$] & Speed [$m/s$] & Classification & Reflection [$-$] & Score \\
		\hline
        2020-09-29T12:19:47.880Z & $51.52132905$ & $5.86072255$ &  $33.41$ & $16.65$ & VEHICLE &   $3.26$ & $0.78$ \\
		2020-09-29T12:19:47.789Z & $51.51840263$ & $5.85496417$ &  $41.51$ &  $4.73$ &   OTHER & $-13.26$ & $0.6$ \\
		2020-09-29T12:19:47.867Z & $51.52171110$ & $5.88431792$ &  $74.63$ & $18.34$ & UNKNOWN & $-36.01$ & $0.35$ \\
		2020-09-29T12:19:46.973Z & $51.52369287$ & $5.85944975$ &  $34.64$ &  $8.01$ & UNKNOWN & $-22.60$ & $0.27$ \\
		2020-09-29T12:19:46.115Z & $51.52206908$ & $5.86990915$ &  $84.86$ &  $4.80$ & UNKNOWN & $-34.85$ & $0.04$ \\
		2020-09-29T12:19:47.962Z & $51.52163624$ & $5.86985159$ &  $51.69$ &  $4.36$ & UNKNOWN & $-33.49$ & $0.04$ \\
		2020-09-29T12:19:48.187Z & $51.52763873$ & $5.86642335$ &  $44.86$ & $11.91$ & UNKNOWN & $-26.50$ & $0.06$ \\
        2020-09-29T12:19:48.331Z & $51.53229705$ & $5.86715643$ & $129.94$ & $12.3$  & UNKNOWN & $-21.93$ & $0.47$ \\
		\bottomrule
	\end{tabular}
\end{table*}
\noindent
\textbf{Dataset Generation.} This procedure includes a summary of the operations performed in the dataset generation. The steps are detailed in Algorithm~\ref{algo:dataset_generation} as following:
\begin{enumerate}
    \item For each scenario, the algorithm starts by loading the log file of the drone and filtering the data considering the sample window of $1$ \emph{sec}, used as the index. Suppose multiple samples are associated with the same timestamp index (i.e. when the difference between each of them is less than $1$~ms). In that case, the algorithm considers the first one appearing in the \ac{CSV} file of the log file being analyzed.
    \item After selecting the row from the log, the algorithm scans the \ac{CSV} file of each radar and \ac{RFDF} system of the same scenario to bind the rows found based on the value of the timestamp index.
    If multiple samples are selected, the algorithm filters consider the closest one to the log's timestamp; otherwise, the algorithm inserts a blank row.
    \item Data enhancement operations are performed on every sample, \emph{i.e.} (i) the conversion of the coordinates from the \ac{GCS} to \ac{UTM} coordinates system, (ii) adding columns with extra data valuable to help training algorithms or (iii) the number of expected drones in the scenario or its classification. Table~\ref{tab:geo2utm} shows an example of coordinates conversion from the \ac{GCS} system to the \ac{UTM} system.
\end{enumerate}
\begin{algorithm}
    \caption{Algorithm for dataset generation.}
    \label{algo:dataset_generation}

    \textbf{Functions:}
    \begin{itemize}
        \item $load\_log\_files(i)$: It loads the log file(s) of $i$-th scenario.
        \item $load\_data(i)$: loads the \ac{CSV} files related to Alvira, Arcus, Venus, and Diana of $i$-th scenario.
        \item $load\_sample\_data\_close\_to\_timestp(k)$: It scans the scenario \ac{CSV} files of Alvira, Arcus, Venus, and Diana and considers, for each of them, the sample with the closest timestamp of the considered one.
        \item $preprocess\_data(k)$: For each element of the $k$-th sample, this function applies the Standard Scaler and the Label Encoder.
        \item $enhancement\_operations(k)$: It enhances the $k$-th sample with extra information (e.g. conversion of coordinates or information related to the scenario).
    \end{itemize}
    \textbf{Inputs:}
     \begin{itemize}
        \item $l_i$: log file(s) of the $i$-th scenario.
        \item $t_i$: sensor Data, coming from Alvira, Arcus, Venus, and Diana, of the $i$-th scenario.
    \end{itemize}
    
    \textbf{Output:}\\
    The dataset $T$ is used to train \ac{ML} models.
    
    \begin{algorithmic}[1]
    \Procedure{DatasetGeneration}{}
    \State $i\gets 0$
    
    \For{$i \gets 0$ to $len(scenarios)$} 
    \State $l_{i} \gets load\_log\_files(i)$
    \State $t_{i} \gets load\_data(i)$ 
    
    \State $k \gets 0$
    \For{$k \gets 0$ to $len(l_{i})$}
    \If{$(timestp_k - timestp_{k+1}) \geq 1 s$}
        \State $load\_sample\_data\_close\_to\_timestp(k)$
        \State $preprocess\_data(k)$
        \State $enhancement\_operations(k)$
    \EndIf
    
    \State $k \gets k + 1$
    \EndFor
    
    \State $i \gets i + 1$
    \EndFor
    \EndProcedure
    \end{algorithmic}
\end{algorithm}

\begin{table}[t]
\caption{Example of coordinates conversion from \ac{GCS} to \ac{UTM}.}
\label{tab:geo2utm}
\centering
\begin{tabular}{ccc}
\toprule
    Format & x                 & y                       \\ \hline
    GCS &      $5.857978$     &       $51.519506$          \\ 
    UTM    & $698276.88$         &   $5711471.08$             \\
\bottomrule
\end{tabular}
\end{table}
The final dataset consists of $5,685$ samples with $57$ columns, used for the dataset analysis described in Section~\ref{sec:ExperimentsAndResults}. Table~\ref{tab:dataset_columns} shows the full list of the columns of the merged dataset, while Table~\ref{tab:dataset_sample} shows a small dataset sample with a subset of its real columns.

\section{Proposed Architecture}
\label{sec:PropArchitecture}

\begin{figure*}[ht]
    \centering
    \includegraphics[width=18cm, height=9cm]{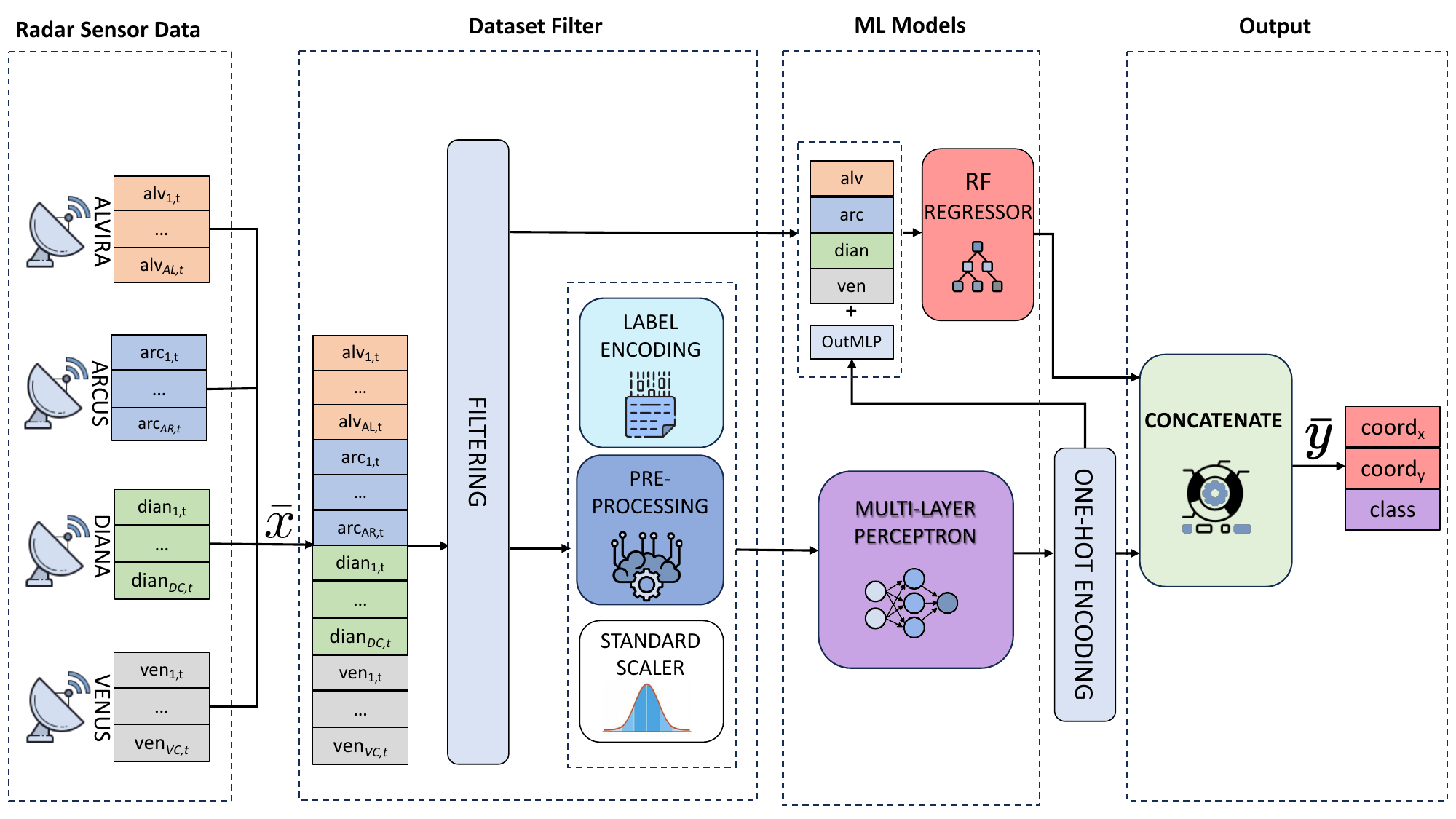}
    \caption{Proposed model architecture.}
    \label{fig:model_schema}
\end{figure*}

\begin{table*}[]
\centering
\caption{Full representation of the columns in the merged dataset. $-$ means that the unit measurement is not specified.}
\label{tab:dataset_columns}
\begin{tabular}{cccc}
\toprule
 {Field Name} & {Data Source} & {Unit Measurement} & {Description}                                                                            \\ \hline
 timestamp                                  & drone, radar and \ac{RFDF} sensors logs                          & $s$                              & \begin{tabular}[c]{@{}c@{}}The reference timestamp\\  of the sample\end{tabular}                                       \\ \hline
scenario\_name                             &                                        & $-$                                  & \begin{tabular}[c]{@{}c@{}}Name of the scenario \\ the sample belongs to\end{tabular}                                  \\
latitude                                   &                                        & decimal degrees                      &                                                                                                                        \\
longitude                                  &                                        & decimal degrees                      &                                                                                                                        \\
altitude                                   &                                        & $m$                               &                                                                                                                        \\ 
speed                                      &                                        & $m/s$                                  & \multirow{-4}{*}{\begin{tabular}[c]{@{}c@{}}Information related to the first \\ drone, always identified\end{tabular}} \\
latitude\_2                                &                                        & decimal degrees                      &                                                                                                                        \\
longitude\_2                               &                                        & decimal degrees                      &                                                                                                                        \\
altitude\_2                                &                                        & $m$                               &                                                                                                                        \\
speed\_2                                   & \multirow{-9}{*}{drone's log}                  & $m/s$                                  & \multirow{-4}{*}{\begin{tabular}[c]{@{}c@{}}Information, related to the second \\ drone, if identified\end{tabular}}   \\ \hline
AlviraPosition\_Latitude                   &                                        & decimal degrees                      &                                                                                                                        \\
AlviraPosition\_Longitude                  &                                        & decimal degrees                      &                                                                                                                        \\
AlviraPosition\_Altitude                   &                                        & $m$                               &                                                                                                                        \\
AlviraVelocity\_Azimuth                    &                                        & degrees                              &                                                                                                                        \\
AlviraVelocity\_Elevation                  &                                        & $m$                               &                                                                                                                        \\
AlviraVelocity\_Speed                      &                                        & $m/s$                                  & \multirow{-6}{*}{\begin{tabular}[c]{@{}c@{}}Information of the identified \\ object by the radar\end{tabular}}         \\
Alvira\_Classification                     &                                        & $-$                                  & \begin{tabular}[c]{@{}c@{}}Classification value of the \\ identified object\end{tabular}                               \\  
Alvira\_Reflection                         &                                        & $-$                                  & Reflection value                                                                                                       \\ 
Alvira\_Score                              &                                        & percentage                           & Score of the produced classification                                                                                   \\ 
Alvira\_Alarm                              & \multirow{-10}{*}{Alvira}              & boolean                              &                                                                                                                        \\ \hline
ArcusPosition\_Latitude                    &                                        & decimal degrees                      &                                                                                                                        \\
ArcusPosition\_Longitude                   &                                        & decimal degrees                      &                                                                                                                        \\ 
ArcusPosition\_Altitude                    &                                        & $m$                               &                                                                                                                        \\ 
ArcusVelocity\_Azimuth                     &                                        & degrees                              &                                                                                                                        \\ 
ArcusVelocity\_Elevation                   &                                        & $m$                               &                                                                                                                        \\ 
ArcusVelocity\_Speed                       &                                        & $m/s$                                  & \multirow{-6}{*}{\begin{tabular}[c]{@{}c@{}}Information of the identified \\ object by the radar\end{tabular}}         \\ 
Arcus\_Classification                      &                                        & $-$                                  & \begin{tabular}[c]{@{}c@{}}Classification value of \\ the identified object\end{tabular}                               \\  
Arcus\_Reflection                          &                                        & $-$                                  & Reflection value                                                                                                                    \\  
Arcus\_Score                               &                                        & percentage                           & Score of the produced classification                                                                                   \\
Arcus\_Alarm                               & \multirow{-10}{*}{Arcus}               & boolean value                        & Alarm value                                                                                                                   \\ \hline
DianaSignal\_snr\_dB                       &                                        & $dB$                                   & SNR of the detected signal                                                                                             \\  
DianaSignal\_bearing\_deg                  &                                        & degrees                              & Antenna's bearing                                                                                                      \\ 
DianaSignal\_range\_m                      &                                        & $m$                               & Antenna's range                                                                                                        \\ 
DianaClassification\_type                  & \multirow{-4}{*}{Diana}                & $-$                                  & Void or controller                                                                                                     \\ \hline
Venus\_isThreat                            &                                        & $-$                                  &                                                                                                                        \\ 
VenusLinkType\_Uplink                      &                                        & $-$                                  & Void or FHSS                                                                                                           \\ 
Venus\_VenusName                           &                                        & $-$                                  & Name of the identified object                                                                                          \\  
Venus\_RadioId                             &                                        & $-$                                  & Id of the identified radio                                                                                             \\ 
Venus\_LifeStatus                          &                                        & $-$                                  & \begin{tabular}[c]{@{}c@{}}Identifies the status of the \\ antenna (down/active)\end{tabular}                          \\  
Venus\_Frequency                           &                                        & $Hz$                                   &                                                                                                                        \\  
Venus\_FrequencyBand                       &                                        & $-$                                  &                                                                                                                        \\  
Venus\_Azimuth                             &                                        & degrees                              & Antenna's azimuth                                                                                                      \\ 
Venus\_Deviation                           & \multirow{-9}{*}{Venus}                &                                      &                                                                                                                        \\ \hline
reference\_classification & Classification algorithm & $-$ & \begin{tabular}[c]{@{}c@{}}Label added during the preprocessing \\phase to train the \ac{MLP} model\end{tabular} \\
\bottomrule
\end{tabular}
\end{table*}
In this work, we propose a real-time \ac{ML} framework called \proto\ to identify, classify and track \acp{UAV}.
As shown in Figure~\ref{fig:model_schema} the input is a vector $\bar{x}$ with the processed information coming from radar and \ac{RFDF} systems. At the same time, the output is the identification, classification, and position of an \ac{UAV}, represented as $\textbf{y} = [class, coord_x, coord_y]$. In particular, the $\textit{class}$ can assume the following values $\{0,1,2,3,4,5,6\}$, where $0$ defines the case in which no drone is inside the \ac{NDZ}, and the remaining values outline the presence of drones. More information are detailed in Section~\ref{sec:ExperimentsAndResults}-\ref{sec:PreliminaryAnalysis}.

In the following sections, we describe each part of the model schema shown in Figure~\ref{fig:model_schema}. Specifically, in Section~\ref{sec:dataset_preparation}, we present dataset preparation steps, while in Section~\ref{sec:ExpSettConfig}, we describe the several components adopted in the framework.

\subsection{Dataset Preparation}
\label{sec:dataset_preparation}

Let us consider as input vector $\textbf{x} = \{ alv_{t} \cup 
arc_{t} \cup dian_{t} \cup ven_{t} \} \in \R^{1,n}$, where $\mathbf{alv_{t}}, \mathbf{arc_{t}}, \mathbf{dian_{t}}, \mathbf{ven_{t}}$ are the vector data of Alvira, Arcus, Diana, and Venus at time $t$, respectively. Specifically, $n$ is the sum of the columns of the vector data provided by the aforementioned sensors. The output vector is $\textbf{y} = [class, coord_x, coord_y] \in R^{1,3}$, where $class \in \N$, $coord_x \in \R$ and $coord_y \in \R$. The parameter $class$ represents the identification and classification of an object, while $coord_x$ and $coord_y$ depict the position of an \ac{UAV}.\\
The features from the sets $x_{1, i}$, $x_{2, i}$, $x_{3, i}$, $x_{4, i}$ where $i$ is the scenario index — sourced from Arcus, Alvira, Diana, and Venus, respectively — are combined with the two log files $l$ namely $l_{1, i}$ (log file of the first drone) and $l_{2, i}$ (log file of the second drone). The goal is to build a consolidated dataset, namely $T$, for training \ac{ML} models. Table~\ref{table:freq_vars} summarises the notations throughout this article.

\begin{table}[!ht]
\centering
\caption{Notation and brief description.}
\label{table:freq_vars}
    \begin{tabular}{ll}
    \toprule
    Notations & Description  \\
    \hline 
    $x$            & The input vector to the framework at time $t$                                                                        \\
    $arc_{t}$      & Vector data coming from Arcus at time $t$                                                                            \\
    $alv_{t}$      & Vector data coming from Alvira at time $t$                                                                           \\
    $dian_{t}$     & Vector data coming from Diana at time $t$                                                                            \\
    $ven_{t}$      & Vector data coming from Venus at time $t$                                                                            \\
    $y$            & \begin{tabular}[c]{@{}l@{}}The output vector of the framework \\ after processing the input vector $x$\end{tabular}  \\
    $class$        & \begin{tabular}[c]{@{}l@{}}Identification and classification value\\  of an object\end{tabular}                      \\
    $coord_x$      & Regressed position of a UAV, $x$ coordinate                                                                          \\
    $coord_y$      & Regressed position of a UAV, $y$ coordinate                                                                          \\
    $AR$           & Number of columns of the CSV files of Arcus                                                                          \\
    $AL$           & Number of columns of the CSV files of Alvira                                                                         \\
    $DC$           & Number of columns of the CSV files of Diana                                                                          \\
    $VC$           & Number of columns of the CSV files of Venus                                                                          \\
    $x_{1,i}$      & Features of Arcus, from $i-th$ scenario                                                                              \\
    $x_{2,i}$      & Features of Alvira, from $i-th$ scenario                                                                             \\
    $x_{3,i}$      & Features of Diana, from $i-th$ scenario                                                                              \\
    $x_{4,i}$      & Features of Venus, from $i-th$ scenario                                                                              \\
    $l_{1,i}$      & Log file of the drone, from $i-th$ scenario                                                                          \\
    $l_{2,i}$      & \begin{tabular}[c]{@{}l@{}}Second log file of the drone, \\ from $i-th$ scenario, if available\end{tabular}          \\
    $T$            & Merged dataset used to train ML models                                                                               \\
    \bottomrule
    \end{tabular}
    \end{table}

\subsection{Model Settings}
\label{sec:ExpSettConfig}
The developed framework is designed to perform multiple operations to identify, track, and classify potential drones.
The main tasks of \proto~performed in \emph{real-time} are:
\begin{itemize}
    \item The framework starts by collecting all the information from the available sensors (both the Radar systems and the \ac{RFDF} systems).
    \item Then \proto\ identifies if there is a drone in the \ac{NDZ} zone, using a binary classification, i.e., \textit{DRONE} - \textit{NO DRONE}. 
    \item When a drone is detected, the framework shows its position on a map and classification information. 
\end{itemize}

\begin{figure}[ht]
    \centering
    \includegraphics[width=10cm, height=5cm]{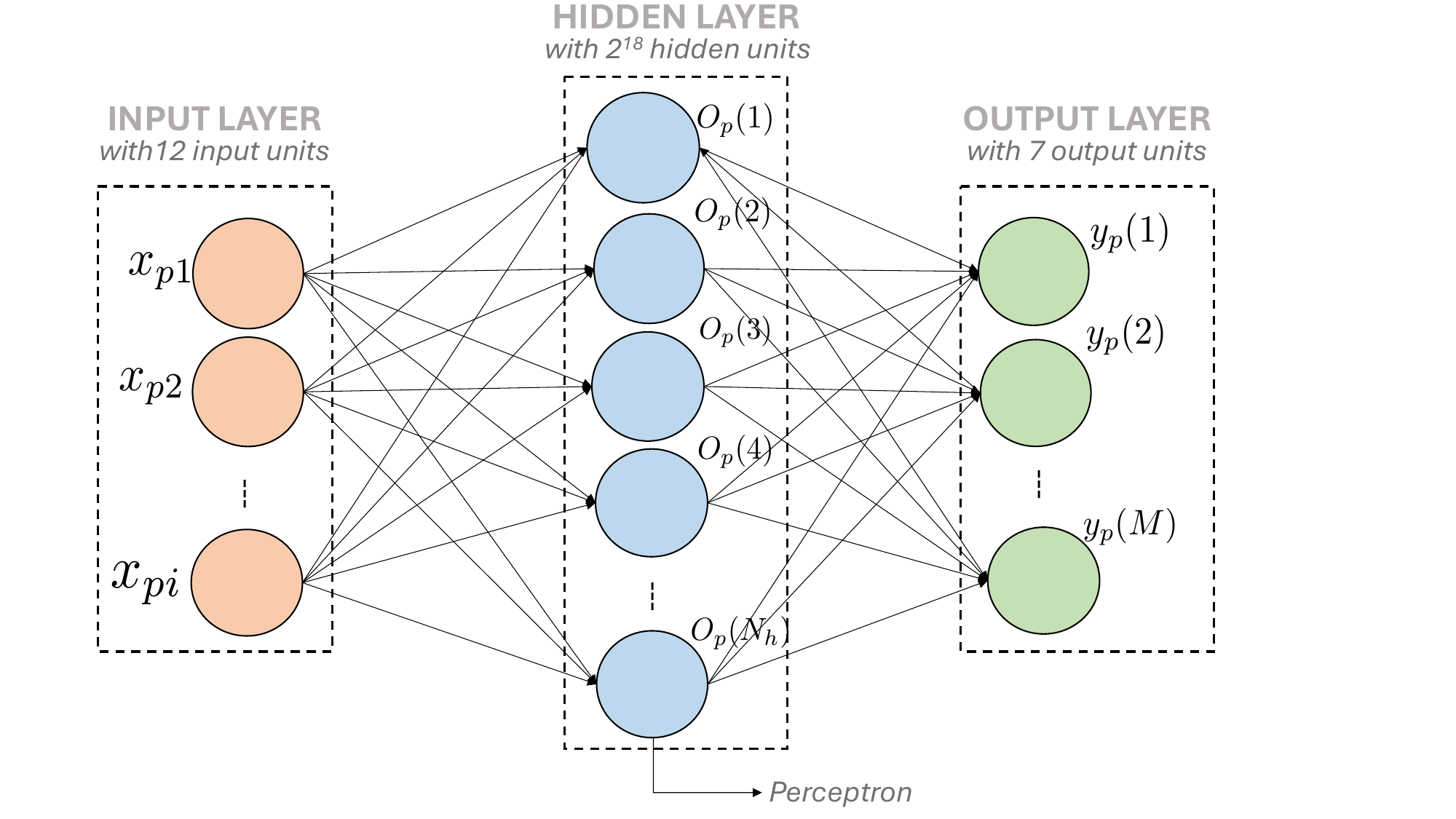}
    \caption{Schema representation of a \acp{MLP} network used in the \proto~framework for classification task. The considered version is structured with $12$ neurons in the input layer, $2^{18}\ (512\times512)$ neurons in the hidden layer(s), and $7$ neurons in the output layer.}
    \label{fig:mlp_schema}
\end{figure}

Moreover, we verify the the design of the proposed \ac{ML} model to achieve a high performance. After evaluating different network configurations, the final setting is the following:

\begin{itemize}
    \item \textit{Radar Sensor Data}: This module communicates with information sources - radars and \ac{RFDF} systems, specifically - to collect data and make it available for subsequent modules.
    
    \item \textit{Dataset Filter}: It is composed of a Label Encoder and a Standard Scaler~\cite{murphy_ml}, as described in Section~\ref{sec:preprocessing_prims}. This module converts the input sensor data to values compatible with the network, for instance, adapting numerical values to be more suitable for \ac{ML} models or converting raw categorical data into numbers. This module is also responsible for selecting specific parts of the input features provided as input to \ac{ML} models according to their needs.
    
    \item \textit{\ac{MLP} Network}: This network performs the identification and classification tasks, returning an output indicating whether a single drone or multiple drones are present within the \ac{NDZ}. The network is defined by $12$ input neurons, $2^{18}\ (512\times512)$ hidden neurons, and $7$ output neurons. Further we adopt the One-Hot encoding~\cite{rodriguez2018beyond} in order to represent the $7$ categorical classes (\texttt{CASE\_UNKNOWN}, \texttt{CASE\_FIXED\_WING}, \texttt{CASE\_MAVIC\_PRO}, \texttt{CASE\_PHANTOM4\_PRO}, \texttt{CASE\_MAVIC2}, \texttt{CASE\_PHANTOM4\_PRO\_MAVIC2}, \texttt{CASE\_PHANTOM4\_PRO\_MAVICPRO}---further details are described in Section~\ref{sec:ExperimentsAndResults}). For example, the scenario that involves the presence of two drones in the \ac{NDZ} (\texttt{CASE\_PHANTOM4\_PRO\_MAVICPRO}) will be encoded with the array $[0,0,0,0,0,1]$.
    
    \item \textit{\ac{RF} Regressor}: This model has $250$ estimators used to predict the position of the identified drones, and it uses a mix of information from the dataset $T$ and the \ac{MLP} network.

    \item \textit{Output}: This module interacts with both the \ac{MLP} Network and the \ac{RF} Regressor, consolidating their outcomes in a single output array. The results are then displayed on the interface, such as on a map.
\end{itemize}

According to the proposed defined model (Fig.~\ref{fig:model_schema}), the \ac{MLP} layer only accepts standardized data (for continuous values) or categorical data expressed using a Label Encoder (for categorical values), whereas the \ac{RF} Classification and Regressor layer accepts raw, unprocessed data.

\section{Experiment and Results}
\label{sec:ExperimentsAndResults}
In this section, we summarize the rationale of the dataset analysis, the metrics and the results, and the implementation details of the \proto\ framework.
In Section~\ref{sec:PreliminaryAnalysis}, we present the preliminary analysis. In this step, we estimate the useful and exploitable information of the dataset $T$ and some details regarding the labelling procedure required to train the considered \ac{ML} models. Next, in Section~\ref{sec:ml_algos}, we evaluate the performance of the \ac{MLP} and \ac{RF} trained models. Finally, we provide the implementation details in Section~\ref{sec:implementation_setup}.

\subsection{Rationale of the Analysis}
\label{sec:PreliminaryAnalysis}

\begin{table*}[t]
\centering
\caption{Study of the useful information exploitable for the \ac{ML} models for each scenario.}
\label{table:dataset_eval}
	\hspace*{0.7cm}
	\begin{tabular}{cccccccc}
        \toprule	
		Scenario & Samples & Drones & No-Drones & Multi-Drone & Samples \% & Drones \% & No-Drones \% \\
		\hline
        scenario 1.1 & $799$  & $126$ & $673$  &  $0$ & $14.05$  & $15.76$ & $84.23$ \\
		scenario 1.2 & $838$  & $233$ & $605$  &  $0$ & $14.74$  & $27.8$  & $72.19$ \\
		scenario 1.3 & $504$  & $162$ & $342$  &  $0$ &  $8.86$  & $32.14$ & $67.85$ \\
		scenario 1.4 & $568$  & $225$ & $343$  &  $0$ &  $9.99$  & $39.61$ & $60.38$ \\
		scenario 2.1 & $1134$ & $372$ & $762$  & $30$ & $19.94$  & $32.8$  & $67.19$ \\
		scenario 2.2 & $1227$ & $671$ & $556$  & $55$ & $21.58$  & $54.68$ & $45.31$ \\
		scenario 3   & $615$  & $107$ & $508$  &  $0$ & $10.81$  & $17.39$ & $82.60$ \\
		\bottomrule
	\end{tabular}
\end{table*}

\begin{table*}[ht]
\centering
\caption{Representation of a portion of the generated dataset as an example with some rows and columns.}
\label{tab:dataset_sample}
\begin{tabular}{llllllll}
\toprule
Timestamp  & Latitude  & Altitude & Speed & Alvira Latitude & Alvira Longitude & Alvira Altitude & Alvira Speed  \\
\hline
$1601456510$ & $51.521736$ & $39.7$     & $13.86$ & $51.52033776$                         & $5.86232428$                           & $32.28467599$                         & $8.54083729$                        \\
$1601456511$ & $51.521832$ & $39.8$     & $14.0$  & $51.52037459$                         & $5.86249421$                           & $33.22264418$                         & $8.82986069$                        \\
$1601456512$ & $51.521912$ & $39.8$     & $13.86$ & $51.5203907$                          & $5.86265874$                           & $34.19893019$                         & $8.95946789$                        \\
$1601456513$ & $51.522006$ & $39.8$     & $13.93$ & $0.0$                                 & $0.0$                                  & $0.0$                                 & $0.0$                               \\
$1601456514$ & $51.522085$ & $39.8$     & $13.93$ & $51.52042446$                         & $5.86282519$                           & $35.14916094$                         & $8.95709419$                        \\
$1601456515$ & $51.522181$ & $39.8$     & $13.86$ & $51.52044779$                         & $5.86299669$                           & $36.16634507$                         & $9.2426815$                         \\
$1601456516$ & $51.522259$ & $39.8$     & $13.86$ & $51.5204803$                          & $5.86317456$                           & $37.17205324$                         & $9.52820396$                        \\
$1601456517$ & $51.522346$ & $39.9$     & $13.86$ & $0.0$                                 & $0.0$                                  & $0.0$                                 & $0.0$                               \\
$1601456518$ & $51.522434$ & $39.7$     & $13.93$ & $51.52052729$                         & $5.86335146$                           & $38.18974604$                         & $9.64759445$                        \\
$1601456519$ & $51.522522$ & $39.7$     & $13.79$ & $51.52054668$                         & $5.86352701$                           & $39.26478952$                         & $9.72277355$                        \\
$1601456520$ & $51.522609$ & $39.8$     & $14.07$ & $51.52055977$                         & $5.86369514$                           & $40.29877538$                         & $9.31249523$          \\
\bottomrule
\end{tabular}
\end{table*}

As depicted in Figure~\ref{fig:scenario}, \ac{RFDF} sensors are positioned close to each other in the center of the test field, while radars are positioned at the edges of the \ac{NDZ}. Radar sensors provide data on the classification, position, bearing (\emph{degrees}), range (\emph{meters}), and reflection of the flying entities. These data are based on the kinematic and reflectivity characteristics of the radar. \ac{RFDF} sensors, on the other side, contribute to identify \acp{UAV} (or the drone controller) based on the \ac{RF} signature.
Before the design phase of \ac{ML} algorithms used in~\proto, a preliminary analysis phase is performed in order to (i) estimate the number of useful information used, (ii) set project parameters, and (iii) split the dataset into subsets (train, validation, and test).

In order to train the \ac{MLP} network using a supervised procedure, the \textit{reference\_classification} column is added to the dataset during the preprocessing phase, as shown in Tab.~\ref{tab:dataset_columns}. The adopted procedure assigns one of the following values for every row of the dataset $T$:

\begin{itemize}
	\item \texttt{CASE\_UNKNOWN}: No drone is identified.
	\item \texttt{CASE\_FIXED\_WING}: The Parrot fixed-wing drone is identified.
	\item \texttt{CASE\_MAVIC\_PRO}: A multi-copter DJI Mavic Pro is identified.
	\item \texttt{CASE\_PHANTOM4\_PRO}: A DJI Phantom 4 is identified.
	\item \texttt{CASE\_MAVIC2}: A DJI MAVIC 2 drone is identified.
	\item \texttt{CASE\_PHANTOM4\_PRO\_MAVIC2}: Both a Phantom 4 Pro and Mavic 2 are identified.
	\item \texttt{CASE\_PHANTOM4\_PRO\_MAVICPRO}: Both a Phantom 4 Pro and a Mavic Pro are identified.
\end{itemize}

The classification highlighted corresponds to the position detected by the radar sensor systems, specifically Alvira and Arcus. If the position of the identified object by the radars is within a deviation of $50$ \emph{meters} from the drone's current position - recorded in the log file(s) - then the procedure assigns to the highlighted row a category ranging from \texttt{CASE\_FIXED\_WING} to \texttt{CASE\_PHANTOM4\_PRO\_MAVICPRO}, depending on the nature of the identified object(s). If not, it assigns \texttt{CASE\_UNKNOWN}.
Specifically, fixed-wing and quadcopters (also known as multi-rotor drones) show specific differences. Indeed, the adopted classification procedures take into account their different properties, such as (i) speed, (ii) flight time, and (iii) manoeuvrability. For instance, fixed-wing drones fly faster than multi-rotor drones, making them ideal for exploring large sites quickly.
Based on these assumptions, it is possible to estimate the number of useful samples to train \ac{ML} algorithms, as shown in Table~\ref{table:dataset_eval}. In particular, the table shows a summary of the information available in the generated dataset $T$, starting from the (i) drone logs, (ii) radar sensor data, and (iii) \ac{RFDF} sensors data.

As the dataset is divided into scenarios and not all samples contain helpful information for training, the amount of useful information is considerably limited. Indeed, on average, only $31\%$ of the dataset contains a real drone position that is useful for our training.

\subsection{Metrics and Results}
\label{sec:ml_algos}
This section describes the training and design procedures to build the modules that process the raw input data. We introduce the metrics and the results for the \ac{MLP} classifier in Section~\ref{sec:mlp_class}, as well as for the \ac{RF} regressor in Section~\ref{sec:rf_regressor}.

\subsubsection{MLP classifier}
\label{sec:mlp_class}
The first network trained during our development phase is the \ac{MLP} classifier. In general, a large amount of data is required to train \acp{MLP} for minimizing the selected loss function and train them on how to generalize from the input dataset. The initial $5,685$ samples of the dataset $T$ are split into three sets: $70\%$ for training, $1.5\%$ for validation and $28.5\%$ for testing. A standardization procedure is employed to enhance the network's training process, and a Label Encoder is used to encode categorical data in numerical values. For the output layer, One-Hot encoding is applied to represent the categories.

The training procedure aims to minimize a loss function evaluated during the training process on the train and validation sets. At the end of the process, the final network performances are evaluated on the test set. For the \ac{MLP} network classifier, the selected loss function is the \emph{Categorical Cross-Entropy Loss}, defined by Eq.~\ref{eq:crossentropy}. The predicted probability distribution for each input is a vector of $C$ values, where $C$ is the number of classes. Each value represents the probability of the input belonging to a specific class. The true probability distribution for each input is also a vector of $C$ values with the same meaning. The categorical cross-entropy loss is defined as:
\begin{equation}\label{eq:crossentropy}
    \text{CE} = -\sum_{i=1}^C y_i\log(\hat{y}_i)
\end{equation}
where $CE$ is the value of the Categorical Cross-Entropy Loss, $y_i$ and $\hat{y}_i$ are the true and predicted probability of the input belonging to class $i$, respectively.

\begin{figure}[htbp]
	\hspace*{-0.2cm}
	\centering
 \includegraphics[width=\columnwidth]{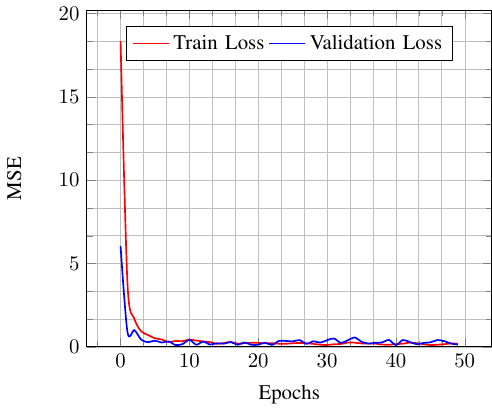}
	\caption{Training and Validation losses for the MLP classifier used for the \emph{Identification} and \emph{Classification} tasks.}
	\label{fig:mlp_training}
\end{figure}

The training procedure is performed during the development phase for $50$ epochs, specifically with a mini-batch size of $5$ and a \emph{Learning Rate} of $0.003$. Further, we adopt the Adam optimizer~\cite{adam_opt} to provide adaptive learning rates for faster and more stable convergence of our \ac{DL} model. Figure~\ref{fig:mlp_training} shows the evolution of the loss value with the epochs. 

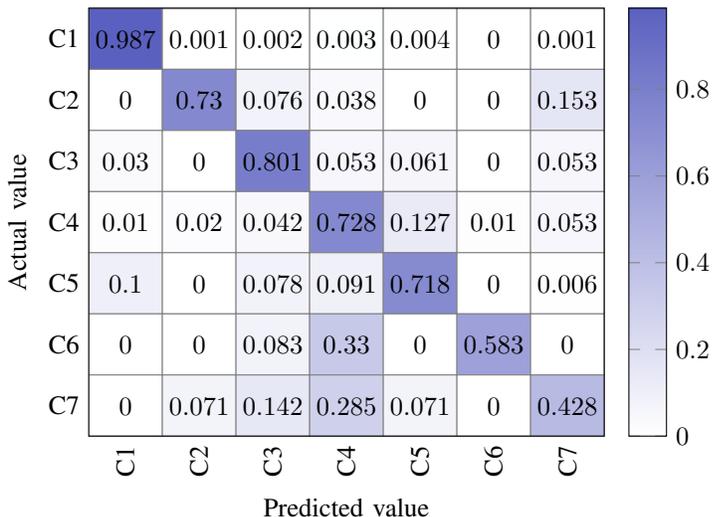
\begin{figure}[h]
    \hspace*{-0.5cm}
    \centering
    \begin{tikzpicture}[scale=1,baseline]
    \pgfkeys{
    /pgf/number format/fixed,
    /pgf/number format/precision=5
    }
	\begin{axis}[
		colormap={bluewhite}{color=(white) rgb255=(90,96,191)},
		xlabel=Predicted value,
		xlabel style={yshift=-5pt},
		ylabel=Actual value,
		ylabel style={yshift=-8pt},
		xticklabels={C1, C2, C3, C4, C5, C6, C7}, 
		xtick={0,...,6}, 
		xtick style={draw=none},
		yticklabels={C1, C2, C3, C4, C5, C6, C7}, 
		ytick={0,...,6}, 
		ytick style={draw=none},
		enlargelimits=false,
		colorbar,
		xticklabel style={
			rotate=90
		},
		nodes near coords={\pgfmathprintnumber\pgfplotspointmeta},
		nodes near coords style={
			yshift=-7pt
		},
		]
		\addplot[
		matrix plot,
		mesh/cols=7,
		point meta=explicit,draw=gray
		] table [meta=C] {
			x y C
			0 0 0.987
			1 0 0.001
			2 0 0.002
			3 0 0.003
			4 0 0.004
			5 0 0
			6 0 0.001
			
			0 1 0
			1 1 0.73
			2 1 0.076
			3 1 0.038
			4 1 0
			5 1 0
			6 1 0.153
			
			0 2 0.03
			1 2 0
			2 2 0.801
			3 2 0.053
			4 2 0.061
			5 2 0
			6 2 0.053
			
			0 3 0.01
			1 3 0.02
			2 3 0.042
			3 3 0.728
			4 3 0.127
			5 3 0.01
			6 3 0.053
			
			0 4 0.1
			1 4 0
			2 4 0.078
			3 4 0.091
			4 4 0.718
			5 4 0
			6 4 0.006

			0 5 0
			1 5 0
			2 5 0.083
			3 5 0.33
			4 5 0
			5 5 0.583
			6 5 0

			0 6 0
			1 6 0.071
			2 6 0.142
			3 6 0.285
			4 6 0.071
			5 6 0
			6 6 0.428
		};
	\end{axis}
    \end{tikzpicture}
    \caption{Confusion matrix of the MLP network classifier.}
    \label{fig:mlp_confusion_matrix}
\end{figure}

The final values of the loss for the last epoch are $0.18$ and $0.13$, for the training and validation phases, respectively. Table~\ref{table:mlp_classifier_perf} shows a summary of the final performances of the \ac{MLP} classifier, while Figure~\ref{fig:mlp_confusion_matrix} shows its Confusion Matrix.
Specifically, the Confusion Matrix is used to evaluate the performance of classification models by summarizing the counts of true and false positives and negatives. The four metrics derived from the confusion matrix are \emph{Precision}, \emph{Recall}, \emph{F1-score}, and the \emph{Support of each class}, represented in Eq.~\ref{eq:precision},\ref{eq:recall}, \ref{eq:F1}, and the number of samples in each class, respectively.

\begin{equation}\label{eq:precision} Precision = \frac{TP}{TP+FP} \end{equation}
\begin{equation}\label{eq:recall} Recall = \frac{TP}{TP+FN} \end{equation}
\begin{equation}\label{eq:F1} F_1 = 2 \times \frac{Precision \times Recall}{Precision + Recall}\end{equation}

In details, the \ac{TP} represents the model accurately predicts the presence of a drone, \ac{FP} provides the inaccurate prediction of a drone's presence, and \ac{FN} depicts an inaccurate prediction of a drone's absence.

\begin{table}[]
\centering
\caption{Summary of the performances of the MLP classifier network.}
\label{table:mlp_classifier_perf}
    \begin{tabular}{ccccc}
    \toprule
    Classes                          & Precision & Recall & F1-Score & Support \\ \hline
    \multicolumn{5}{c}{\textbf{One-vs-All}}\\
    $0$                                & $0.98$      & $0.99$   & $0.98$     & $1,096$    \\
    $1$                                & $0.73$      & $0.73$   & $0.73$     & $26$    \\
    $2$                                & $0.79$      & $0.8$    & $0.8$      & $131$     \\
    $3$                                & $0.8$       & $0.73$   & $0.76$     & $188$     \\
    $4$                                & $0.74$      & $0.72$   & $0.73$     & $153$     \\
    $5$                                & $0.78$      & $0.58$   & $0.67$     & $12$       \\
    $6$                                & $0.21$      & $0.43$   & $0.28$     & $14$      \\ \hline
    \multicolumn{5}{c}{\textbf{Overall}}\\
    \multicolumn{1}{l}{Accuracy}       & $-$         & $-$      & $0.9$      & $1,620$    \\
    \multicolumn{1}{l}{Macro Avg}      & $0.72$      & $0.71$   & $0.71$     & $1,620$    \\
    \multicolumn{1}{l}{Weighted Avg}   & $0.91$      & $0.9$    & $0.91$     & $1,620$    \\ 
    \bottomrule
    \end{tabular}
\end{table}

\subsubsection{RF regressor}
\label{sec:rf_regressor}
Considering the \ac{RF} network, we filtered the main dataset $T$ by obtaining a sub-dataset with $1,343$ samples. The $80\%$ ($1,074$ samples) are used for the training phase, and the $20\%$ ($269$ samples) for the test phase. It is worth noticing that we do not adopt any standardization procedure in this case, unlike what is done for the \ac{MLP} classifier. We use the \emph{Label Encoder} only for the categorical input columns in the derived subset of the initial dataset to manage them as numbers.

The final training results for the \ac{RF} regressor are evaluated by using the \ac{MSE}, \ac{MAE}, and $R^2$ metrics as depicted in Eq.~\ref{eq:MSE},~\ref{eq:MAE} and~\ref{eq:R2}, respectively:

\begin{equation}\label{eq:MSE} MSE=\frac{1}{n}\sum_{i=1}^{n}E[(\hat{y}_i-y_i)^2] \end{equation}

\begin{equation}\label{eq:MAE} MAE=\frac{1}{n}\sum_{i=1}^{n}|\hat{y}_i-y_i| \end{equation}

\begin{equation}\label{eq:R2} R^2 = 1 - \frac{\sum_{i=1}^{n} (y_i - \hat{y_i})^2}{\sum_{i=1}^{n} (y_i - \bar{y})^2} \end{equation}\\

where $n$ is the number of data points, $y_i$ is the $i$-th observed value of the dependent variable, $\hat{y_i}$ is the $i$-th predicted value of the dependent variable, and $\bar{y}$ is the mean value of the dependent variable across all observations.

The $MSE$, $MAE$, and $R^2$ amount to $0.29$, $0.04$, $0.93$, respectively. The results confirm the robustness and validity of the chosen model to perform this regression task.

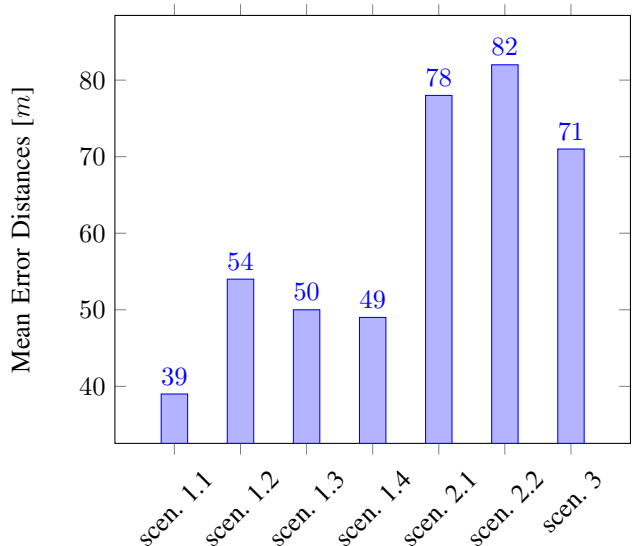
\begin{figure}[htbp]
    \hspace*{-0.5cm}
    \begin{tikzpicture}  
    \begin{axis}  
    [  
        ybar,  
        enlargelimits=0.15,  
        ylabel={Mean Error Distances [$m$]}, 
        symbolic x coords={scen. 1.1, scen. 1.2, scen. 1.3, scen. 1.4, scen. 2.1, scen. 2.2, scen. 3}, 
        xtick=data,  
         nodes near coords, 
        nodes near coords align={vertical}, 
        xticklabel style={
    			rotate=45
    		}
        ]  
    \addplot coordinates {
            (scen. 1.1, 39) 
            (scen. 1.2, 54) 
            (scen. 1.3, 50) 
            (scen. 1.4, 49) 
            (scen. 2.1, 78) 
            (scen. 2.2, 82) 
            (scen. 3, 71) 
        };  
      
    \end{axis}  
    \end{tikzpicture}

    \caption{Mean differences (in meters) between predicted and real drone positions in the various scenarios.}
	\label{fig:med_distances}
\end{figure}

Further, to assess the performance of the regressor, the trained model uses each scenario of the dataset $T$ as input to estimate the medium difference between the real positions of the drone(s) and the predicted ones. Figure~\ref{fig:med_distances} shows the mean differences between predicted and real drone positions in each scenario. In all the considered scenarios, the mean regression error is below $100$ meters, and the models perform better when there is only one drone in the scenario. This behaviour is highly predictable, given the low number of sensors available for training the models and detecting drones.

\subsection{Setup and Implementation}
\label{sec:implementation_setup}

The development and test phases are performed on a custom desktop machine running Arch Linux with the Linux Kernel $6.2.11$ and CUDA $12.1$\cite{cuda_link}. The hardware includes an AMD Ryzen $7$ $2700X$ Eight-Core Processor with $32$ GB of RAM and an NVIDIA $1080$ GPU.

For the \ac{ML} models design and implementation, we developed the solution in Python $3$. In particular, we adopted the following libraries:

\begin{itemize}
    \item \textit{PyTorch} 1.13.1~\cite{pyqt_site}: machine learning framework used for developing and training neural network-based deep learning models in Python. We used this library to model the MLP network.
    \item \emph{sklearn} 1.2.2~\cite{sklearn_site}: for the Standard Scaler and Label Encoder, the train/test dataset split and the Random Forest algorithm implementation.
    \item \emph{numpy} 1.24.2: a common Python library useful to work with arrays and matrices.
    \item \emph{pandas} 2.0~\cite{pandas_site}: a common Python library, particularly for the DataFrame object.
    \item \emph{matplotlib} 3.7.2: used to show and save graphs.
    \item \emph{PyQt} 6.5~\cite{pyqt_site}: the binding of the Qt libraries for Python. This library has been used to create the application's main \ac{GUI}, exploiting its cross-platform nature.
    \item \emph{Leaflet} 1.9.4~\cite{leaflet_site}: the Javascript \ac{GIS} library used to show real and predicted objects on the interactive map of the base.
\end{itemize}

\begin{figure}[htbp]
	\hspace*{-0.2cm}
	\centering
 \includegraphics[width=\columnwidth]{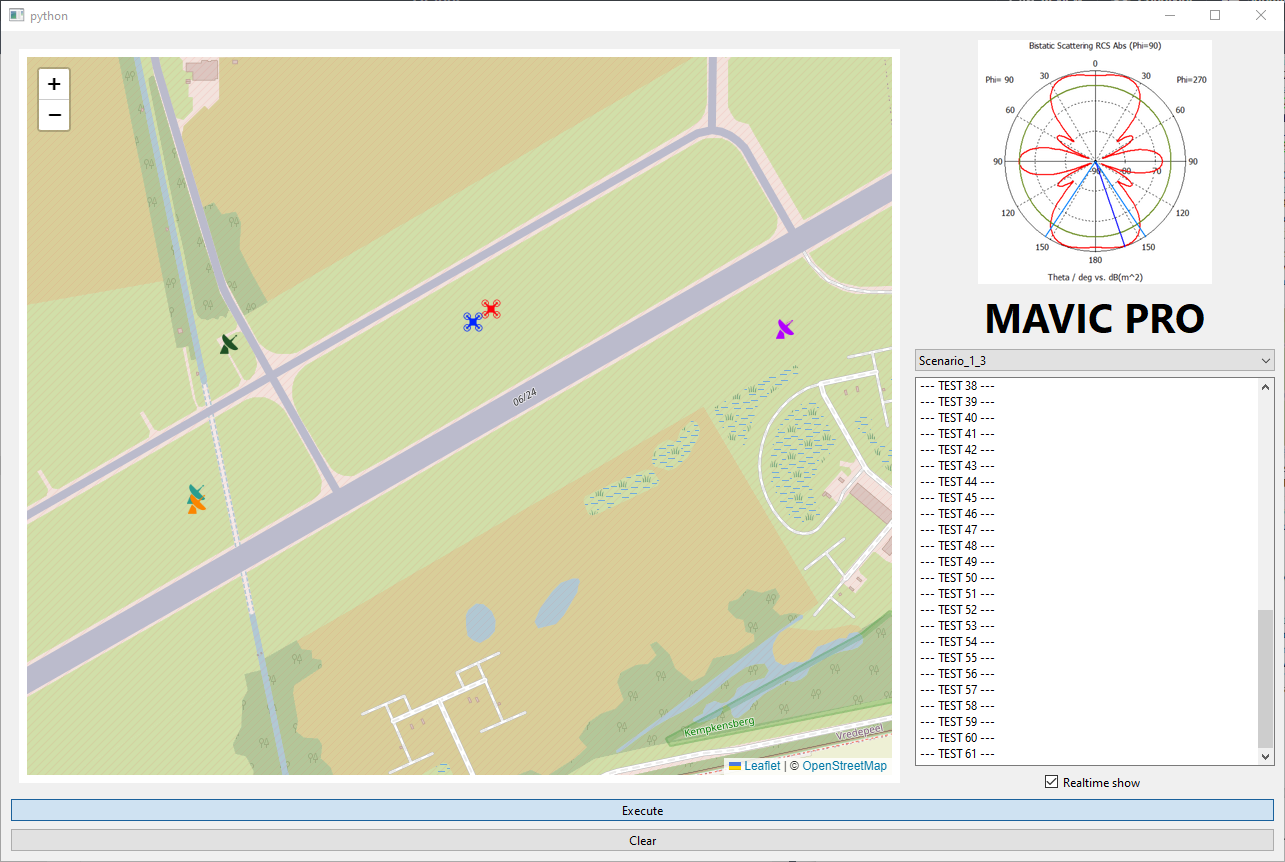}
	\caption{Screenshot of the proposed framework in real-time mode. The blue icon depicts the real position of the drone, while the red icon represents its predicted position. Meanwhile, the other radar icons depict the static positions of both radars and \ac{RFDF} sensors.}
	\label{fig:mainapp_screen}
\end{figure}

Figure~\ref{fig:mainapp_screen} depicts the main functionalities of the \proto~ framework, such as drone(s) identification, classification, and real-time tracking on the map.

\section{Related Work}
\label{sec:RelatedWork}

In this section, we review the state of the art on the Radio Frequency (RF) machine learning and deep learning approaches adopted to detect, identify and track drones. 
Table~\ref{tab:comparison} summarizes the information of related literature and presents the main features of these methods.

\begin{table*}[htbp]
\small
\centering
    \caption{Comparison and overview of related contributions on drones RF identification, classification, and tracking using Machine Learning and Deep Learning techniques. A \checkmark\ symbol indicates the fulfillment of a particular feature, and a \xmark\ symbol denotes the miss of the feature or that the feature is not applicable.}
    \label{tab:comparison}
\resizebox{\textwidth}{!}{
\begin{tabular}{l p{1.3cm} p{1.6cm} p{2.1cm} p{1.6cm} p{1.2cm} p{1.4cm} p{1.5cm} p{1.0cm}}
\toprule
Ref. & Analysis on Dataset & Identification and Classification & ML/DL Model & Real Time & Multiple Drone Detection & Fixed-wing Drone Detection & Path-Tracking & Open Source Code \\ \hline
        \cite{alsad2019_fgcs} & \xmark & \checkmark & DNN & \xmark  & \checkmark & \xmark & \xmark & \checkmark \\ \hline
        \cite{sanjoy2022_tccn} & \xmark & \checkmark & YOLO-lite & \xmark & \xmark & \xmark & \xmark & \xmark \\ \hline
        \cite{alemadi2020_iciot} & \checkmark & \checkmark & CNN & \xmark &  \xmark & \xmark & \xmark & \xmark \\ \hline
        \cite{allahham2020_iciot} & \checkmark & \checkmark & 1D-CNN & \xmark & \xmark & \xmark & \xmark & \xmark \\ \hline
        \cite{medaiyese2021_iconsonics} & \checkmark & \checkmark & XGBoost & \xmark & \xmark & \xmark & \xmark & \xmark \\ \hline
        \cite{sazdicjotic2022_esa} & \checkmark & \checkmark & FC-DNN & \xmark   & \checkmark & \xmark & \xmark & \xmark \\ \hline
        \cite{sanjoy2021_comsnets} & \checkmark & \checkmark & DRNN & \xmark & \checkmark & \xmark & \xmark & \xmark \\ \hline
        \cite{ibrahim2021_sensors} & \checkmark & \checkmark & KNN and XGBoost & \xmark & \xmark & \xmark & \xmark & \xmark \\ \hline
        \cite{wei2021_ieeesj} & \xmark & \checkmark & KNN, SVM and RF & \xmark & \xmark & \xmark & \xmark & \xmark \\ \hline
        \textbf{\proto} & \cmark & \cmark & MLP + RF & \cmark & \cmark& \cmark & \cmark & \cmark \\ \bottomrule
\end{tabular}
}
\end{table*}

For instance, Al-Sa’d \emph{et al.}~\cite{alsad2019_fgcs} collected, analysed, and recorded raw RF signals from several types of drones in different states. Furthermore, they leveraged a deep learning technique to detect and identify malicious drones and their flight mode. The authors designed three \acp{DNN} to (i) detect the drone, (ii) detect the drone and recognize its type, and (iii) detect the drone, and recognize its type and its state. The authors do not consider fixed-wing drones, and they do not perform any path-tracking operation.

Basak \emph{et al.}~\cite{sanjoy2022_tccn} focused on the development of (i) RF drone signal detection, (ii) spectrum localization, and (iii) drone classification by using a two-stage technique. In the first stage, they adopt the \ac{GoF}  sensing for drone detection and the \ac{DRNN} framework for drone classification. In the second stage, they use the \ac{YOLO-lite} framework to perform the combined drone RF signal detection, spectrum localization, and drone classification. However, neither multiple detections of drones nor trajectory tracking on a map are considered.

Al-Emadi \emph{et al.}~\cite{alemadi2020_iciot} proposed a real-time RF drone detection and identification framework to inspect the radio spectrum between the drone and its controller. The solution adopts a \ac{CNN} to train and test an RF dataset released by~\cite{alsad2019_fgcs}. The experimental results show the effectiveness and feasibility of using RF signals in combination with a CNN to detect and identify a drone. The proposed solution achieves an F1 score of $99.7$\% for drone identification. 
Nevertheless, the authors do not consider fixed-wing drones and drone path-tracking operations.

Allahham \emph{et al.}~\cite{allahham2020_iciot} investigated deep learning techniques to perform (i) drone detection, (ii) drone detection and type identification, and (iii) drone detection, type and state identification by using a three multi-channel 1-dimensional \ac{CNN}. The dataset adopted in the experiments is \textit{Drone RF} dataset~\cite{Allahham2019_DRONERF}. The performance for (i) shows an average accuracy of $100$\%, while (ii) has an accuracy of $94.6$\%, and, finally, the last one (iii) presents an accuracy of $87.4$\%.

The authors in~\cite{medaiyese2021_iconsonics} developed an RF machine-learning drone detection and identification system by analyzing the low-band RF signals emitted by the flight controller. They proposed three machine learning models based on \ac{XGBoost} algorithm to detect and identify (i) the presence of a drone, (ii) the presence of a drone and type, and (iii) the presence of a drone, type and the operational mode. The accuracy achieved by the three models is $99.96$\%, $90.73$\%, and $70.09$\%, respectively. The higher the model complexity, the lower the model accuracy. This latter implies the low effectiveness of using the frequency components of a signal as a signature to detect the activities performed by drones. 
From the results achieved by the models, we deduced that using the frequency components of a signal as a signature to detect drone activities is not very effective. Trajectory tracking on a map is not considered in this case.

Sazdić-Jotić \emph{et al.}~\cite{sazdicjotic2022_esa} proposed RF detection and identification algorithms to detect and identify single or multiple drones. They built an RF dataset by considering scenarios with (i) a single drone, (ii) two drones, and (iii) three drones. They detect and identify a single drone with an accuracy of $99.8$\% and $96.1$\%, respectively, while the results of detecting multiple drones show an average accuracy of $97.3$\%. 
The deep learning algorithms used are mainly \ac{FC-DNN}. Although the approach performs well, the authors do not consider path tracking.

The authors in~\cite{sanjoy2021_comsnets} presented a \ac{DRNN} that classifies different drone signals in single-drone and multiple-drone scenarios. The authors built an RF dataset with nine commercial drone types, and further, they evaluated the proposed model in \ac{AWGN} and multipath conditions. The model achieved roughly $99$\% classification accuracy for single and simultaneous multi-drone scenarios. However, 
The described approach does not take into account drone path tracking and fixed-wing drones. 

Ibrahim \emph{et al.}~\cite{ibrahim2021_sensors} presented a UAV identification and hierarchical detection approach by leveraging an ensemble learning based on \ac{KNN} and \ac{XGBoost}. The proposed solution can (i) check the availability of a UAV, (ii) specify the type of the UAV, and (iii) determine the flight mode of the detected UAV. This approach reaches a classification accuracy of around $99$\%. However, the authors do not consider drone path tracking.

Wei \emph{et al.}~\cite{wei2021_ieeesj}, proposed a drone detection and identification system based on WiFi signals and high-frequency RF fingerprints.
The system (i) performs UAV detection, (ii) extracts the features \ac{FD}, (iii) \ac{AIB} and \ac{SIB} (iii) adopts the \ac{PCA} algorithm for the feature dimensionality reduction, and (iv) applies the \ac{NCA} algorithm for metric learning. Finally, the authors test \ac{KNN}, \ac{SVM}, and \ac{RF} to identify UAVs. They verified their model in two different scenarios, i.e., indoor with a \ac{SNR} of $10$~dB and outdoor with a \ac{SNR} of $3$~dB. In the indoor scenario, the average identification accuracy of \ac{FD}, \ac{AIB}, \ac{SIB} is $100$\%, $97.23$\%, and $96.11$\% respectively. In the outdoor scenario, the identification accuracy of the same features is $100$\%, $95.00$\%, and $93.50$\%, respectively. The authors do not consider drone trajectory tracking.

To sum up, the discussion above confirms that despite there are several contributions to the state-of-the-art, none of them analyze and evaluate the proposed techniques on both fixed-wing and multi-copter drones for the (i) detection, (ii) classification and (iii) simultaneous tracking, as well as the drone path tracking. Such constraints make previous solutions unsuitable for this problem and call for new domain-specific approaches.
Moreover, none of the approaches in the current literature perform tracking, identification, and classification in real-time, but only offline.

\section{Conclusion}
\label{sec:Conclusion}
In this paper, we proposed \proto, a framework to prevent and detect unauthorized \acp{UAV} for Critical Infrastructures. 
\proto\ can identify, classify, and track multi-copter and fixed-wing drones in real time. Our solution leverages two components: (i) a network of \acl{RFDF} radar sensor network distributed in the No-Drone Zone, and (ii) \acl{C-UAS}, a system adopted to collect and process the data generated by the radar sensor network and detect the presence of any unauthorized or malicious drone. \proto\ features several properties such as: (i) it relies only on the wireless data collected from the \ac{RFDF} sensor network; (ii) it can be extended to detect, classify and track different aerial vehicles at the same time; and (iii) it can be integrated with pre-existing drone detection solutions in compliance with the existing regulations. At the same time, our model has been trained on a dataset comprising UAV flights provided by \ac{NATO}~\cite{Kaggle2021_DATNATO}. Our results show that the trained models achieve a good accuracy of $90$\% for the identification and classification tasks, and we can discriminate between \acp{UAV} and fixed-wings. The \ac{MLP} model achieves an accuracy of $90$\% with a True Positive Rate (Recall) of $\approx 0.71$, and a True Negative Rate of $\approx 0.98$. The \ac{RF} model achieves a \ac{MSE} $\approx 0.29$, \ac{MAE} $\approx 0.04$, $R^2 \approx 0.93$ on the final dataset.
Finally, we highlight that we also released the source code of \proto\ as open-source to foster the replicability of our results, encourage the deployment and extension of \proto, and check the viability of further research directions.

\section*{Acknowledgment}
We would like to acknowledge the use of the “Drone Detection” data source provided by the NATO Communications \& Information Agency (NCIA). The findings reported here are solely the responsibility of the authors.  This work was partially funded by the Italian MISE FSC 2014/20 Asse I project ‘Casa delle Tecnologie Emergenti’, and by the Italian P.O. Puglia FESR 2014/20 project 6ESURE5 ‘Secure Safe Apulia’.

\ifCLASSOPTIONcaptionsoff
  \newpage
\fi

\bibliographystyle{IEEEtran}
\bibliography{biblio}

\end{document}